\documentclass[10pt]{article}
\usepackage[margin=1in]{geometry}
\usepackage[utf8]{inputenc}
\usepackage[T1]{fontenc}
\usepackage[hidelinks]{hyperref}
\usepackage{url}
\usepackage{booktabs}
\usepackage{amsfonts}
\usepackage{nicefrac}
\usepackage{microtype}
\usepackage{graphicx}
\usepackage{bm}
\usepackage{bbm}
\usepackage{siunitx}
\usepackage{amsmath,amsfonts}
\usepackage[numbers,sort&compress]{natbib}
\usepackage[labelfont=bf]{caption}
\usepackage{cleveref}
\usepackage{titling}

\DeclareMathOperator{\E}{\mathbb{E}}

\title{A comparison of approaches to improve worst-case predictive model performance over patient subpopulations}

\author{
    Stephen R. Pfohl$^{*,1}$, Haoran Zhang$^{2}$, Yizhe Xu$^1$, \\ 
    Agata Foryciarz$^{1,3}$, Marzyeh Ghassemi$^{4,5}$, Nigam H. Shah$^1$
}

\date{
    $^1$Stanford Center for Biomedical Informatics Research, Stanford University, Stanford, California 94305, USA \\
    $^2$Department of Computer Science, University of Toronto, Toronto, Ontario, Canada \\
    $^3$Department of Computer Science, Stanford University, Stanford, California 94305, USA \\
    $^4$Department of Electrical Engineering and Computer Science, Massachusetts Institute of Technology, Cambridge, Massachusetts 02139, USA \\
    $^5$Institute for Medical and Evaluative Sciences, Massachusetts Institute of Technology, Cambridge, Massachusetts 02139, USA
    \\
    *Correspondence to: \texttt{spfohl@stanford.edu}
}

\begin{document}
\maketitle

\begin{abstract}
    Predictive models for clinical outcomes that are accurate on average in a patient population may underperform drastically for some subpopulations, potentially introducing or reinforcing inequities in care access and quality. Model training approaches that aim to maximize worst-case model performance across subpopulations, such as distributionally robust optimization (DRO), attempt to address this problem without introducing additional harms. We conduct a large-scale empirical study of DRO and several variations of standard learning procedures to identify approaches for model development and selection that consistently improve disaggregated and worst-case performance over subpopulations compared to standard approaches for learning predictive models from electronic health records data. In the course of our evaluation, we introduce an extension to DRO approaches that allows for specification of the metric used to assess worst-case performance. We conduct the analysis for models that predict in-hospital mortality, prolonged length of stay, and 30-day readmission for inpatient admissions, and predict in-hospital mortality using intensive care data. We find that, with relatively few exceptions, no approach performs better, for each patient subpopulation examined, than standard learning procedures using the entire training dataset. These results imply that when it is of interest to improve model performance for patient subpopulations beyond what can be achieved with standard practices, it may be necessary to do so via data collection techniques that increase the effective sample size or reduce the level of noise in the prediction problem.
\end{abstract}

\section{Introduction}

Predictive models learned from electronic health records are often used to guide clinical decision-making.
When patient-level risk stratification is the basis for providing care, the use of models that fail to predict outcomes correctly for one or more patient subpopulations may introduce or perpetuate inequities in care access and quality \cite{rajkomar2018-ji,chen2020ethical}.
Therefore, the assessment of differences in model performance metrics across groups of patients is among an emerging set of best practices to assess the ``fairness'' of machine learning applications in healthcare \cite{chen2019can,coley2021racial,Seyyed-Kalantari2020,park2021comparison,barda2021addressing,pfohl2019creating,zink2020fair}.
Other best practices include the use of participatory design and transparent model reporting, including critical assessment of the assumptions and values embedded in data collection and in the formulation of the prediction task, as well as evaluation of the benefit that a model confers given the intervention that it informs \cite{chen2020ethical,obermeyer2019dissecting,benjamin2019assessing,paulus2020predictably,Vyas2020,jacobs2021measurement,Passi2019,sendak2020presenting,gebru2018datasheets,Mitchell2019,sorelle_2021_impossibility,jung2021framework}.

One approach for addressing fairness concerns is to declare \textit{fairness constraints} and specify a constrained or regularized optimization problem that encodes the desire to predict an outcome of interest as well as possible while minimizing differences in a model performance metric or in the distribution of predictions across patient subpopulations \cite{Hardt2016,Agarwal2018,Celis2018,Zafar2019-cx}.
A known concern with this approach is that it often does not improve the model for \textit{any} group and can reduce the fit of the model or induce miscalibration for \textit{all} groups, including the ones for whom an unconstrained model performed poorly, due to differences in the distribution of the data collected for those subpopulations that limit the best-achievable values for the metric of interest \cite{Kleinberg2016,Chouldechova2017,barocas-hardt-narayanan,Pfohl_2021,martinez2020minimax,Liu2019}.
Furthermore, satisfying such constraints does not necessarily promote fair decision-making or equitable resource allocation \cite{liu2018delayed,hu2020fair,fazelpour2020algorithmic,Corbett-Davies2018}.

As an alternative to equalizing model performance across groups of patients, recent works have proposed maximizing \textit{worst-case} performance across pre-defined subpopulations, as a form of \textit{minimax fairness} \cite{martinez2020minimax,diana2021minimax,Sagawa*2020Distributionally}, representing a shift in perspective towards the goal of identifying the best model for each patient subpopulation.
The objective of this work is to compare approaches formulated to improve disaggregated and worst-case model performance over subpopulations -- through modifications to training objectives, sampling approaches, or model selection criteria -- with standard approaches to learn predictive models from electronic health records.
We evaluate multiple approaches for learning predictive models for several outcomes derived from electronic health records databases in a large-scale empirical study.
In these experiments, we define patient subpopulations in terms of discrete demographic attributes, including racial and ethnic categories, sex, and age groups.
We compare empirical risk minimization (ERM; the standard learning paradigm) applied to the entire training dataset with four alternatives: (1) training a separate model for each subpopulation, (2) balancing the dataset so that the amount of data from each subpopulation is equalized, (3) model selection criteria that select for the best worst-case performance over subpopulations, and (4) distributionally robust optimization (DRO) approaches \cite{ben2013robust,Hu2018,Sagawa*2020Distributionally} that directly specify training objectives to maximize a flexible notion of worst-case performance over subpopulations.
We evaluate each of these approaches in terms of their capability to improve several model performance metrics overall, for each subpopulation, and in the worst-case over subpopulation compared to ERM applied to the entire training dataset.

\section{Results}

\begin{table}[!t]
\centering
\caption{Summary of prediction tasks across databases and outcomes}
\label{tab:cohort_summary}
\begin{tabular}{llll}
\toprule
Database  & Outcome                & Summary statistics & Reference      \\ \midrule
STARR     & In-hospital mortality          & Table \ref{tab:cohort_starr_admissions} & \citet{Pfohl_2021} \\
STARR     & Hospital LOS \textgreater{}= 7 days     &  Table \ref{tab:cohort_starr_admissions} & \citet{Pfohl_2021} \\
STARR     & 30-day readmission             & Table \ref{tab:cohort_starr_admissions} &  \citet{Pfohl_2021} \\
MIMIC-III & In-hospital mortality &  Supplementary Table \ref{tab:cohort_mimic_eicu} & \citet{harutyunyan2019multitask} \\
eICU      & In-hospital mortality          &  Supplementary Table \ref{tab:cohort_mimic_eicu} & \citet{sheikhalishahi2020benchmarking} \\
\bottomrule
\end{tabular}
\end{table}

\subsection{Cohort characteristics}

\begin{table}[!th]
\centering
\caption{Characteristics of the inpatient admission cohort drawn from the STARR database. Data are grouped based on age, sex, and race/ethnicity. Further context regarding the operationalization of race and ethnicity is included in the methods section. Shown, for each group, is the number of patients extracted and the incidence of in-hospital mortality, prolonged length of stay (LOS), and 30-day readmission.}
\label{tab:cohort_starr_admissions}
\begin{tabular}{lrrrr}
\toprule
{} & {} & \multicolumn{3}{c}{Outcome Incidence} \\
\cmidrule{3-5}
Group &    Count &  In-hospital mortality &  Prolonged LOS &  30-day readmission \\
\midrule
\lbrack18-30)  &   24,638 &              0.00690 &                     0.174 &              0.0455 \\
\lbrack30-45)  &   47,177 &             0.00613 &                     0.129 &               0.0390 \\
\lbrack45-55)  &   28,847 &              0.0179 &                     0.208 &              0.0527 \\
\lbrack55-65)  &   37,717 &              0.0251 &                     0.229 &              0.0556 \\
\lbrack65-75)  &   38,555 &              0.0291 &                     0.238 &              0.0563 \\
\lbrack75-90)  &   35,206 &              0.0408 &                     0.239 &              0.0555 \\
\midrule
Female   &  120,677 &              0.0162 &                     0.166 &              0.0453 \\
Male     &   91,455 &              0.0275 &                     0.246 &              0.0572 \\
\midrule
Asian    &   30,551 &              0.0217 &                     0.176 &               0.054 \\
Black or African American  &    8,189 &              0.0199 &                     0.242 &              0.0602 \\
Hispanic or Latino &   37,299 &              0.0186 &                     0.197 &              0.0534 \\
Other race/ethnicity  &   24,649 &              0.0294 &                     0.205 &              0.0431 \\
White    &  111,452 &              0.0201 &                     0.205 &              0.0494 \\
\bottomrule
\end{tabular}
\end{table}

We define five prediction tasks across three electronic health records databases and three outcomes (Table \ref{tab:cohort_summary}), structured in two categories: (1) the prediction of in-hospital mortality, prolonged length of stay, and 30-day readmission upon admission to the hospital and (2) the prediction of in-hospital mortality during the course of a stay in the intensive care unit (ICU).
These tasks are selected for consistency with prior published work \cite{Pfohl_2021,harutyunyan2019multitask,sheikhalishahi2020benchmarking} and to enable the examination of the generalizability of results across a diverse set of databases containing structured longitudinal electronic health records and temporally-dense intensive care data.

We directly follow \citet{Pfohl_2021} to create cohorts from the STARR \cite{Datta2020} database for learning models that predict in-hospital mortality, prolonged length of stay (hospital length of stay greater than or equal to seven days), and 30-day readmission upon admission to the hospital. 
This cohort consists of 212,140 patients, and is slightly larger than in \citet{Pfohl_2021} due to ongoing refresh of the STARR database (Table \ref{tab:cohort_starr_admissions}).
We extract cohorts from the MIMIC-III \cite{johnson2016mimic} and eICU \cite{pollard2018eicu} databases for learning models that predict in-hospital mortality using data collected in intensive care settings using the definitions from two recent benchmarking studies \cite{harutyunyan2019multitask,sheikhalishahi2020benchmarking}. 
The cohorts extracted from the MIMIC-III and eICU databases contain 21,139 and 30,680 patients, respectively (Supplementary Table \ref{tab:cohort_mimic_eicu}).

\subsection{Experimental overview}
\begin{figure}[!t]
    \centering
    \includegraphics[width=\linewidth]{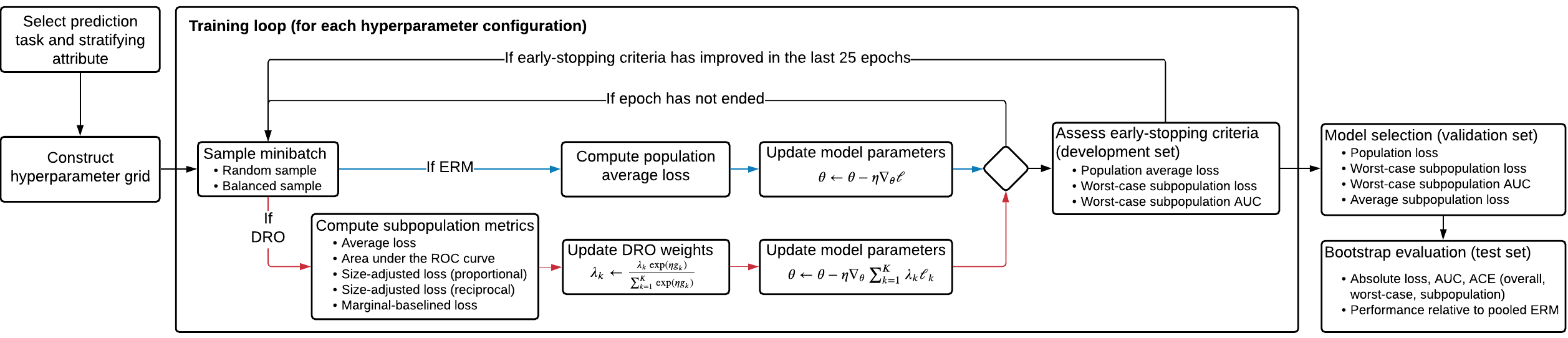}
    \caption{A schematic representation of the experimental procedure. 
    Prior to the execution of the experiments, we extract, for each prediction task, clinical data elements recorded prior to the occurrence of a task-specific index event, which defines the portion of a patient's longitudinal record that can be used as inputs to predictive models (fully-connected feed-forward networks, gated recurrent units (GRUs) \cite{cho2014_gru}, and logistic regression).
    For each prediction task and stratifying attribute, we evaluate each element of a hyperparameter grid that includes hyperparameters related to the choice of model class, training objective, sampling rule, and early-stopping stopping criteria.
    Following training, we evaluate several model selection criteria and evaluate the selected models on a held-out test set. 
    }
    \label{fig:training_loop}
\end{figure}

Figure \ref{fig:training_loop} provides an overview of the experimental procedure and further details are provided in the Methods section. 
For each prediction task, we learn a model using standard training and model selection approaches as a baseline.
These models are learned with ERM applied to the entire training dataset (pooled ERM). 
This approach relies on stochastic gradient descent applied in a minibatch setting, where each batch is randomly sampled from the population without regards to subpopulation membership, and training terminates via an early-stopping rule that assesses whether the average population cross-entropy loss, has failed to improve, consecutively over a fixed number of iterations, on a held-out development set. 
Model selection is by a grid search to identify the hyperparameters that minimize the population average loss on a held-out validation set.

For each combination of prediction task and stratifying attribute (race and ethnicity, sex, and age group), we conduct comparisons with several alternative configurations of ERM, as described in section \ref{sec:methods_training_model_selection}. 
The first alternative that we consider is one where the standard training and model selection approaches are applied separately for each subpopulation (stratified ERM).
Then, we evaluate, in isolation and composition, modifications both to the sampling and early-stopping approaches used during training and to the model selection criteria applied over the hyperparameter grid search.
The modified sampling rule is such that each minibatch seen during training is balanced to have an equal proportion of samples from each subpopulation during training, similar to sampling approaches taken in imbalanced learning settings \cite{he2009learning}.
We further evaluate worst-case early-stopping approaches that are based on identifying the model with the lowest worst-case loss or largest worst-case area under the receiver operating characteristic curve (AUC) over subpopulations during training.
We evaluate the worst-case early-stopping rules in conjunction with worst-case model selection criteria that select hyperparameters based on the best worst-case performance on a held-out validation set.
We report on the results for models selected based on the worst-case model selection over a combined grid over model-class-specific hyperparameters, the sampling rule, and the early-stopping criteria.

In addition to variations of ERM, we evaluate several variations of DRO (section \ref{sec:dro_methods}). 
Each DRO approach can be interpreted as ERM applied to the distribution with the worst-case model performance under a class of distribution shifts.
By casting the class of distribution shifts in terms of \textit{subpopulation shift}, \textit{i.e.} shifts in the subpopulation composition of the population, the training objective becomes aligned with maximizing worst-case performance across subpopulations.
Each of the DRO approaches that we assess corresponds to a different way of assessing relative model performance across subpopulations.
We use the unadjusted formulation of \citet{Sagawa*2020Distributionally} to define model performance for each subpopulation in terms of the average cross-entropy loss.
As comparisons of the loss across subpopulations may not be contextually meaningful in cases when differences in the outcome incidence are present, we evaluate additive adjustments to the loss (section \ref{sec:additive_adjustment}) that scale with the estimated negative marginal entropy of the outcome (the \textit{marginal-baselined loss}).
We also evaluate additive adjustments that scale with the relative size of the subpopulation, either proportionally \cite{Sagawa*2020Distributionally} or inversely, to account for differences in the rate of overfitting that may result due to differences in the sample size.
We further propose an alternative DRO formulation that allows for flexible specification of the metric used to define worst-case performance (section \ref{sec:alternative_metrics}).  
In our experiments, we evaluate this formulation using comparisons of the AUC across subpopulations to define worst-case performance.
As in the case of ERM, we evaluate DRO approaches over a hyperparameter grid that includes balanced and unbalanced sampling rules, early stopping criteria, and objective-specific hyperparameters, but report only the results that follow from the application of the two worst-case model selection criteria (loss and AUC), separately for each of the five DRO configurations and in the aggregate over all DRO configurations.

After model selection, we assess overall, disaggregated, and worst-case model performance on a held-out test set in terms of the AUC, the average loss, and the absolute calibration error (ACE) \cite{Pfohl_2021,Austin2019,Yadlowsky2019}.
Confidence intervals for the value of each metric are constructed via the percentile bootstrap with 1,000 bootstrap samples of the test set. 
Confidence intervals for the relative performance compared to the pooled ERM approach are constructed via computing the difference in each performance metric on each bootstrap sample.

\subsection{Experimental results}

\begin{figure}[!th]
    \centering
    \includegraphics[width=0.9\linewidth]{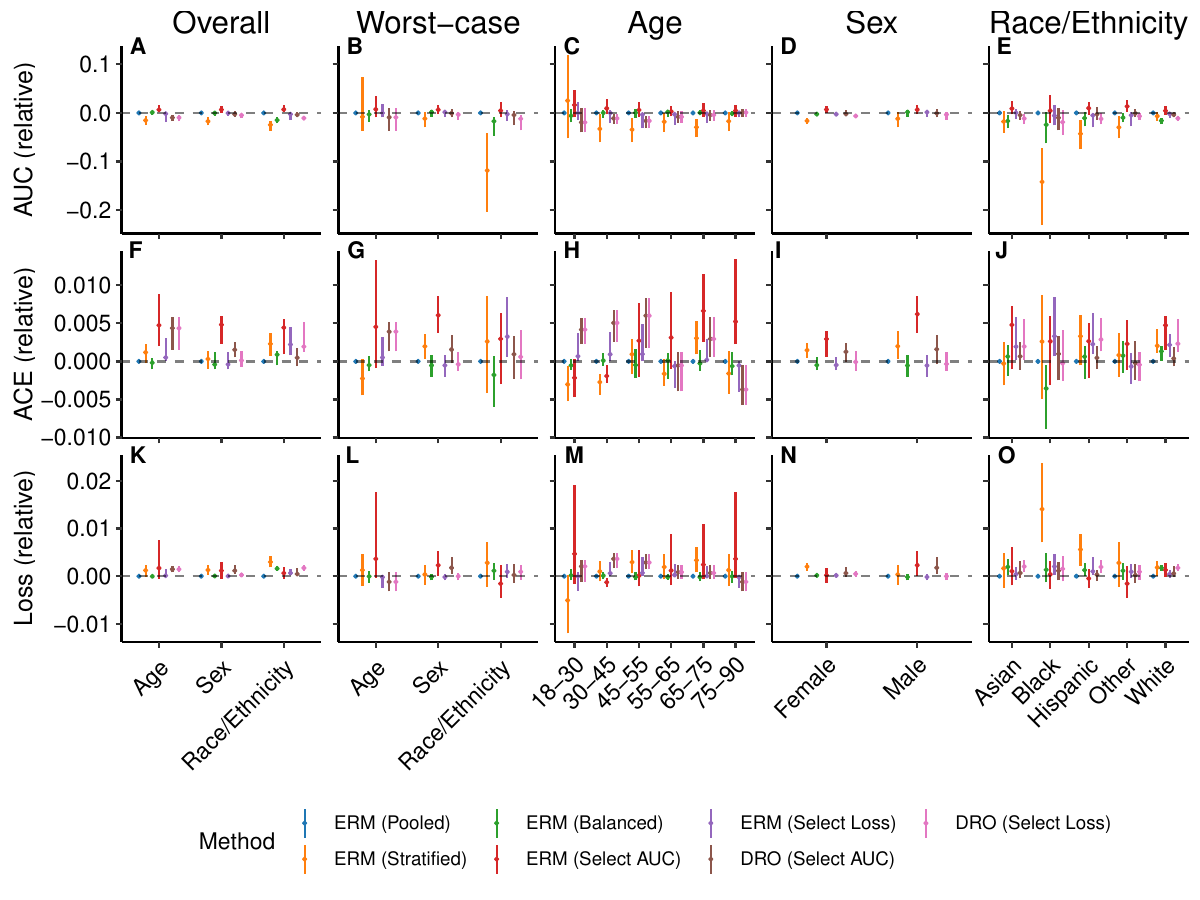}
    \caption{
    The performance of models that predict in-hospital mortality at admission using data derived from the STARR database.
    Results shown are the area under the receiver operating characteristic curve (AUC), the absolute calibration error (ACE), and the loss assessed in the overall population, on each subpopulation, and in the worst-case over subpopulations for models trained with pooled, stratified, and balanced ERM and a range of distributionally robust optimization (DRO) training objectives, relative to the results attained by applying empirical risk minimization (ERM) to the entire training dataset.
    For both pooled ERM and DRO, we show the models selected based on worst-case model selection criteria that perform selection based on the worst-case subpopulation AUC (Select AUC) or loss (Select Loss).
    Model selection occurs over all relevant training objectives, sampling rules, and early-stopping criteria.
    Error bars indicate 95\% confidence intervals derived with the percentile bootstrap with 1,000 iterations.
    }
    \label{fig:starr_mortality_performance_relative}
\end{figure}

\begin{figure}[!ht]
    \centering
    \includegraphics[width=0.9\linewidth]{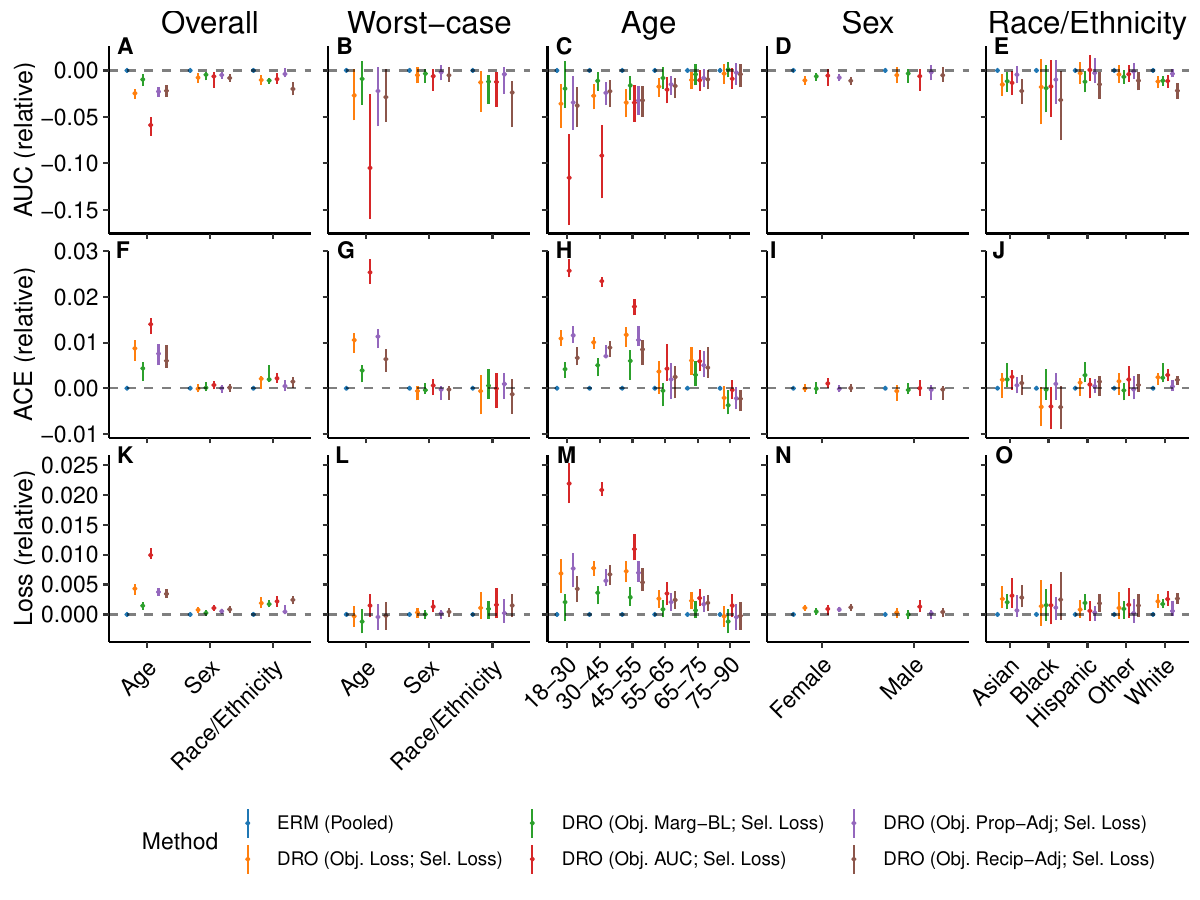}
    \caption{
    The performance of models trained with distributionally robust optimization (DRO) training objectives to predict in-hospital mortality at admission using data derived from the STARR database, following model selection based on worst-case loss over subpopulations.
    Results shown are the area under the receiver operating characteristic curve (AUC), absolute calibration error (ACE), and the loss assessed in the overall population, on each subpopulation, and in the worst-case over subpopulations for models trained with the unadjusted DRO training objective (Obj. Loss), the adjusted training objective that subtracts the marginal entropy in the outcome (Obj. Marg-BL), the training objective that uses the AUC-based update (Obj. AUC), and training objectives that use adjustments that scale proportionally (Obj. Prop-Adj) and inversely to the size of the group (Obj. Recip-Adj), relative to the results attained by applying empirical risk minimization (ERM) to the entire training dataset.
    Error bars indicate 95\% confidence intervals derived with the percentile bootstrap with 1,000 iterations.
    }
    \label{fig:starr_mortality_dro_loss_relative}
\end{figure}
In the main text, we primarily report results for all approaches examined relative to the results attained by applying empirical risk minimization to the entire population (pooled ERM).
We report detailed findings for models that predict \textit{in-hospital mortality} using data drawn from the STARR database.
In the supplementary material, we report absolute and relative performance metrics for models derived from all cohorts and prediction tasks.

The model that predicts in-hospital mortality using data drawn from the STARR database attains an AUC of 0.827, 95\% CI [0.81, 0.83], an ACE of 0.0027, 95\% CI [0.0012, 0.0035], and a loss of 0.090, 95\% CI [0.088, 0.091].
We observe differences in the performance characteristics of models learned with pooled ERM across subpopulations defined by stratification on age, sex, and race and ethnicity (Supplementary Figure \ref{fig:starr_mortality_performance}).
The observed subpopulation losses for the pooled ERM model are ordered on the basis of the incidence of the outcome, with few exceptions (Supplementary Figure \ref{fig:starr_mortality_performance}W,X,Y). 
We observe relatively little variability in AUC when stratifying by race and ethnicity (AUC [95\% CI]: 0.84 [0.81, 0.86], 0.82 [0.77, 0.87], 0.84 [0.82, 0.87], 0.84 [0.81, 0.86], 0.80 [0.79, 0.82] for the Asian, Black, Hispanic, Other, and White subpopulations, respectively; Supplementary Figure \ref{fig:starr_mortality_performance}E), but do observe differences when stratifying by sex (AUC [95\% CI]: 0.85 [0.83, 0.86] and 0.78 [0.77, 0.86] for the female and male subpopulations, respectively; Supplementary Figure \ref{fig:starr_mortality_performance}D) and by age group (AUC [95\% CI]: 0.73 [0.64, 0.80], 0.89 [0.86, 0.92], 0.82 [0.78, 0.85], 0.82 [0.80, 0.84], 0.79 [0.76, 0.81], 0.73 [0.70, 0.75] for the 18-30, 30-45, 45-55, 55-65, 65-75, and 75-90 age groups, respectively; Supplementary Figure \ref{fig:starr_mortality_performance}C).
While the model is well-calibrated overall, we observe poorer calibration for the Black subpopulation (ACE [95\% CI]: 0.0065 [0.0043, 0.011]) and for the youngest (0.0063 [0.0053, 0.0076]) and oldest (0.0068 [0.0035, 0.010]) subpopulations (Supplementary Figure \ref{fig:starr_mortality_performance}).

With few exceptions, the approaches assessed did not improve on the models for in-hospital mortality trained with pooled ERM using the STARR database, in terms of performance metrics assessed overall, in the worst-case, and on each subpopulation (Figure \ref{fig:starr_mortality_performance_relative}).
We observe that balanced sampling and stratified training approaches generally did not improve performance, except for improvements in calibration for some cases: balanced sampling improved calibration for the Black population (change in ACE [95\% CI]: -0.0035 [-0.0089, -0.00048]; Figure \ref{fig:starr_mortality_performance_relative}J) and stratified training improved calibration for the 18-30 and 30-45 age groups (-0.0030 [-0.0052, -0.00058] and -0.0027 [-0.0044, -0.0016], respectively; Figure \ref{fig:starr_mortality_performance_relative}H).
Model selection based on the worst-case AUC over subpopulations improved the overall AUC (change in overall AUC [95\% CI]: 0.0067 [0.0012, 0.016], 0.0067 [0.0083, 0.014], 0.0072 [0.0013, 0.016] for stratification based on age, sex, and race and ethnicity, respectively; Figure \ref{fig:starr_mortality_performance_relative}A), but these improvements were not reflected in improvements in worst-case or subpopulation AUC, with the exception of an improvement in the AUC for patients in the ``Other'' race and ethnicity category (change in AUC [95\% CI]: 0.013 [0.0025, 0.027]; Figure \ref{fig:starr_mortality_performance_relative}E) and an improvement in AUC for the female population (change in AUC [95\% CI]: 0.0070 [0.00019, 0.015]; Figure \ref{fig:starr_mortality_performance_relative}I).
Furthermore, model selection on the basis of the worst-case AUC criteria increased overall calibration error (Figure \ref{fig:starr_mortality_performance_relative}F) and failed to improve the calibration error or the loss for any subpopulation, with the exception of the patients in the 30-45 age group (Figure \ref{fig:starr_mortality_performance_relative}H,M).

DRO approaches to learning models to predict in-hospital mortality from data in the STARR database did not generally improve on models built with pooled ERM. 
The only exception is that the models selected on the either the worst-case loss or AUC across age groups led to a minor improvement in calibration error for the 75-90 age group (change in ACE [95\% CI]: -0.0037 [-0.0057, -0.00045]; Figure \ref{fig:starr_mortality_performance_relative}H). 
Furthermore, when stratifying by sex or race and ethnicity, the DRO variants performed similarly, regardless of whether the worst-case loss or AUC was used for model selection (Figure \ref{fig:starr_mortality_dro_loss_relative}A,B,D,E,F,G,I,J,K,L,N,O and Supplementary Figures \ref{fig:starr_mortality_dro_loss},\ref{fig:starr_mortality_dro_auc}).
When stratifying by age group, we observe increased calibration error and loss and reduced AUC, particularly for younger age groups, the magnitude of which differ substantially across DRO approaches, with the models trained with the AUC-based DRO objective showing the largest reduction in performance and those trained with the marginal-baselined approach showing the smallest (Figure \ref{fig:starr_mortality_dro_loss_relative}C,H,M).

For the remainder of the cohorts and prediction tasks, pooled ERM performed the best overall, in the worst-case, and for each subpopulation assessed, with few exceptions. 
For models that predict \textit{prolonged length of stay} using the STARR database, we observe improvements in overall calibration, without improvements in loss, for stratified ERM and some instances of DRO, when age group or race and ethnicity is used for stratification (Supplementary Figures \ref{fig:starr_los_performance},\ref{fig:starr_los_dro_loss},\ref{fig:starr_los_dro_auc}).
For models that predict \textit{30-day readmission} from the data in the STARR database, we observe no improvements relative to pooled ERM (Supplementary Figures \ref{fig:starr_readmission_performance},\ref{fig:starr_readmission_dro_loss},\ref{fig:starr_readmission_dro_auc}).
Among models that predict \textit{in-hospital mortality} from intensive care databases, following \citet{harutyunyan2019multitask} and \citet{sheikhalishahi2020benchmarking}, those trained with pooled ERM perform best overall, in the worst-case, and for each subpopulation (Supplementary Figures \labelcref{fig:mimic_mortality_vitals_performance,fig:mimic_mortality_vitals_dro_loss,fig:mimic_mortality_vitals_dro_auc,fig:eicu_mortality_performance,fig:eicu_mortality_dro_loss,fig:eicu_mortality_dro_auc}).
In some cases, we observe large degrees of variability in the performance estimates, likely as a result of the small size of the subpopulations examined (\textit{e.g.} when assessing AUC for the 18-30 population drawn from MIMIC-III; Supplementary Figures \ref{fig:mimic_mortality_vitals_performance},\ref{fig:mimic_mortality_vitals_dro_loss},\ref{fig:mimic_mortality_vitals_dro_auc}).

\section{Discussion}

Our experiments provide a large-scale empirical evaluation of approaches formulated to improve disaggregated and worst-case performance across subpopulations.
In summary, none of the approaches evaluated consistently improved overall, worst-case, or disaggregated model performance compared to models learned with ERM applied to the entire training dataset. 
Our empirical findings parallel recent theoretical and other empirical results that demonstrate the limitations of approaches enabling robustness under distribution shift and generalization out-of-distribution \cite{rosenfeld2021the,rosenfeld2021online,Koh2021,zhang2021empirical,gulrajani2020search,taori2020measuring}.
The presence of situations where at least one alternative approach improved model performance for at least one subpopulation compared to ERM applied to the entire training dataset suggests that it may be worthwhile to routinely evaluate these approaches to identify the set of the subpopulation-specific models with the highest performance, but our results do not provide clear insight into when and if those approaches should be preferred.
Our results suggest that the alternative ERM approaches, \textit{i.e.} those that use stratified training, balanced subpopulation sampling, or worst-case model selection, typically outperform the DRO approaches without incurring the additional computational burden of tuning DRO-specific hyperparameters.

A limitation of our experiments is that we primarily evaluate high-capacity models learned from large datasets with subpopulation structure defined based on a single demographic attribute.
This may mask potential benefits that may be present only when learning with lower-capacity models, from smaller cohorts, or in the presence of extreme imbalance in the amount of data from each subpopulation. 
The existence of such benefits would mirror the results of experiments demonstrating the efficacy of self-supervised pre-training in improving accuracy of predictive models learned from small cohorts \cite{McDermott2021,steinberg2021language}.
A further implication of considering only a single stratifying attribute is that it has the potential to mask \textit{hidden stratification}, \textit{i.e.} differences in model properties for unlabeled subpopulations or for intersectional ones defined across attributes \cite{Oakden-Rayner2020-wc}.
Introducing a larger space of discrete groups via the intersection of a pre-defined set of attributes is a straightforward approach that may help alleviate this concern, although it also leads to a combinatorial increase in the number of subpopulations and a reduction in sample size for each subpopulation.
However, even with the current experimental procedure, we observe imprecise estimates of model performance and potentially a lack of power to detect differences in model performance due to the small sample size and event rates for the evaluated subpopulations.
Approaches to combat these issues include sample splitting approaches such as nested cross validation, the incorporation of an auxiliary model into the DRO training objective that learns to identify latent subpopulations for which the model performs poorly, either as a function of multiple attributes or directly from the space of features used for prediction \cite{Sohoni2020,Lahoti2020-py,pmlr-v80-hebert-johnson18a,Kim2018,kearns2017,creager2021}, and the use of model-based estimates of subpopulation performance metrics to increase the sample-efficiency of performance estimates and statistical power of comparisons across small subpopulations \cite{miller2021model}.

A challenge central to this work is the task of defining a well-motivated notion of worst-case performance.
The definition of the worst-case is complicated by the presence of differences in the distribution of the data across subpopulations that affect the best-achievable value of a chosen performance metric.
For example, in our study, we observed substantial differences in the average cross entropy loss observed across subpopulations ordered on the basis of differences in incidence of the outcome that are further essentially unrelated to the ordering of the observed AUC or calibration error across those subpopulations.
Such effects are not unique properties of the average loss, as performance metrics assessed at a threshold, such as the true and false positive rates and the positive predictive value, are also influenced by event rates if calibration is maintained \cite{simoiu2017problem,corbett2017algorithmic,bakalar2021fairness,foryciarz2021evaluating}.
Furthermore, the effect of stratification on the observed AUC can be complex when the stratifying attribute is predictive of the outcome.
The subpopulation AUC reflects the extent to which the model ranks patients belonging to the subpopulation for whom the outcome is observed above those for whom it is not observed, but does not reflect the accuracy of such ranking between patient subpopulations \cite{kallus2019fairness,narasimhan2020pairwise}.

A strength of this work is the flexibility of the notion of worst-case performance considered.
The motivation for the marginal-baselined loss was to adjust the average cross-entropy loss used to assess worst-case performance for differences in the incidence of the outcome by subtracting the entropy attributable to the incidence.
We further introduced a class of DRO training objectives that allow for customization of the metric used to assess worst-case performance (equation (\ref{eq:DRO_lambda_g}).
Here, we used that formulation to reason about worst-case performance in terms of the subpopulation AUC (equation \ref{eq:g_AUC})). 
This approach differs from related works that propose robust optimization training objectives over a broad class of performance metrics \cite{cotter2019optimization,narasimhan2020pairwise} in that we use the AUC only as a heuristic to assess the relative performance of the model across subpopulations in the update over the weights on the subpopulation losses, rather than as the primary objective function over the model parameters. 
A limitation of approaches that directly use the AUC in the update over the model parameters is that they are unlikely to produce calibrated models because direct AUC-maximization only encodes the only encodes the desire to improve ranking accuracy without regards to the calibration of the resulting model.
An interesting future direction is to consider an approach that incorporates a calibration metric into the formulation of equation ($\ref{eq:DRO_lambda_g}$) in order to reduce worst-case miscalibration across subpopulations during training, similar to post-processing approaches formulated for the same purpose \cite{pmlr-v80-hebert-johnson18a,wald2021calibration}.

\subsection{Conclusion}
In this work, in the context of predictive models learned from electronic health records data, we characterized the empirical behavior of model development approaches designed to improve worst-case and disaggregated performance of models across patient subpopulations.
The results indicate that, in most cases, models learned with empirical risk minimization using the entire training dataset perform best overall and for each subpopulation. 
When it is of interest to improve model performance for specific patient subpopulations beyond what can be achieved with this standard practice for a fixed dataset, it may be necessary to increase the available sample size for those subpopulations or to use targeted data collection techniques to identify and collect auxiliary features that reduce the level of noise in the prediction problem \cite{ChenJohanssonSontag_NIPS18}.
In cases where it is of interest to increase the sample size, decentralized aggregation techniques \cite{xu2021federated} as well as large-scale pre-training and transfer learning \cite{McDermott2021,steinberg2021language} may be effective.
Our results do not confirm that applying empirical risk minimization to large training datasets is sufficient for developing equitable predictive models, but rather suggest only that approaches designed to improve worst-case and disaggregated model performance across subpopulations are unlikely to do so in practice.
We emphasize that using a predictive model for allocation of a clinical intervention in a manner that promotes fairness and health equity requires reasoning about the values and potential biases embedded in the problem formulation, data collection, and measurement processes, as well as contextualization of model performance in terms of the downstream harms and benefits of the intervention.

\section{Methods}

\subsection{Cohorts}

\subsubsection{Databases} \label{sec:databases}
\paragraph{STARR}
The Stanford Medicine Research Data Repository (STARR) \citep{Datta2020} is a clinical data warehouse containing deidentified records from approximately three million patients from Stanford Hospitals and Clinics and the Lucile Packard Children's Hospital.
This database contains structured diagnoses, procedures, medications, laboratory tests, vital signs mapped to the Observational Medical Outcomes Partnership (OMOP) Common Data Model (CDM) version 5.3.1, sourced from inpatient and outpatient clinical encounters that occurred between 1990 and 2021.
In this work, we consider data derived from encounters occurring prior to January 30, 2021.
The use of this data was conducted in accordance with all relevant guidelines and regulations.
Approval for the use of STARR for this study is granted by the Stanford Institutional Review Board Administrative Panel on Human Subjects in Medical Research (IRB 8 - OHRP \#00006208, protocol \#57916), with a waiver of informed consent.

\paragraph{MIMIC-III}
The Medical Information Mart for Intensive Care-III (MIMIC-III) database is a publicly and freely available database that consists of deidentified electronic health records for 38,597 adult patients admitted to the intensive care unit of the Beth Israel Deaconess Medical Center between 2001 and 2012 \cite{johnson2016mimic}.
As described in \citet{johnson2016mimic}, this database was created and made available via the Physionet \cite{goldberger2000physiobank} platform following approval by the Massachusetts Institute of Technology Institutional Review Board, with a waiver of informed consent, in accordance with all relevant guidelines and regulations.

\paragraph{The eICU Collaborative Research Database}
The eICU Collaborative Research Database (eICU; Version 2.0) is a publicly and freely available multicenter database containing deidentified records for over 200,000 patients admitted to ICUs across the United States from 2014 to 2015 \cite{pollard2018eicu}.
This data is made available subject to same approvals and access mechanisms as MIMIC-III.

\subsubsection{Cohort definitions}

\paragraph{In-hospital mortality, prolonged length of stay, and 30-day readmission among inpatient admissions in STARR}
We replicate the logic of \citet{Pfohl_2021} to extract a cohort of inpatient admissions and associated outcomes for in-hospital mortality, prolonged length of stay (defined as a hospital length of stay greater than or equal to seven days), and 30-day readmission (defined as a subsequent admission within thirty days of discharge of the considered admission) from the STARR database.
We extract all inpatient hospital admissions spanning two distinct calendar dates for which patients were 18 years of age or older at the date of admission and randomly sample one admission per patient.
The index date is considered to be the date of admission such that only historical data collected prior to admission is used for prediction.

\paragraph{In-hospital mortality in publicly available intensive care databases}
We apply the logic presented in \citet{harutyunyan2019multitask} and \citet{sheikhalishahi2020benchmarking} to extract cohorts from MIMIC-III and eICU appropriate for developing models to predict in-hospital mortality using data collected from the first 48 hours of a patient's ICU stay.
Both cohorts are restricted to patients between 18 and 89 years or age, and exclude admissions that contain more than one ICU stay or an ICU stay shorter than 48 hours.

\subsubsection{Subpopulation definitions} \label{sec:subpopulation}
We define discrete subpopulations based on demographic attributes: (1) a combined race and ethnicity variable based on self-reported racial and ethnic categories, (2) sex, and (3) age at the index date, discretized into 18-30, 30-45, 45-55, 55-65, 65-75, 75-90 years, with intervals exclusive of the upper bound.
Patients whose sex is not recorded as male or female are excluded when sex is considered as the stratifying attribute, and included otherwise.

For cohorts extracted from STARR, we construct a combined race and ethnicity attribute by assigning ``Hispanic or Latino'' if the ethnicity is recorded as ``Hispanic or Latino'', and the value of the recorded racial category otherwise.
The racial categories provided by the upper-level of the OMOP CDM vocabulary correspond to the Office of Management and Budget categories \cite{Ulmer2009}: ``Asian'', ``American Indian or Alaska Native'', ``Black or African American'', ``Native Hawaiian or Other Pacific Islander'', and ``White''. 
Due to limited sample size in some groups, we use an ``Other race/ethnicity'' category that includes ``American Indian or Alaska Native'', ``Native Hawaiian or Other Pacific Islander'', ``Other or no matching race/ethnicity'', ``Patient declined or refused to state'', and ''Unknown race/ethnicity.
Disaggregated statistics associated with these groups are provided in Supplementary Table \ref{tab:cohort_starr_admissions_other_race_eth}.
For succinctness in the presentation of results, we use the following categories: ``Asian'', ``Black'', ``Hispanic'', ``Other'', and ``White''.

For cohorts derived from the MIMIC-III and eICU databases, we map the semi-structured ``Ethnicity'' field provided in those databases to
the following categories: ``Black or African American'', ``White'', and ``Other race/ethnicity''.
In the MIMIC-III database, the ``Other race/ethnicity'' category includes categories that map to ``Asian'', ``Hispanic or Latino'', ``Other or no matching race/ethnicity'', ``Patient refused or declined to state'', and ``Unknown race/ethnicity''. 
For the eICU database, the ``Other race/ethnicity'' category includes ``Asian'', ``Hispanic or Latino'', and ``Other or unknown race/ethnicity''.
Disaggregated statistics associated with these groups are provided in Supplementary Table \ref{tab:cohort_mimic_eicu_other_race_eth}.

\subsection{Feature extraction}
For the cohorts derived from STARR, we apply a procedure similar to the one described in \citet{Pfohl_2021} to extract a set of clinical features to use as input to fully-connected feedforward neural networks and logistic regression models.
The features are based on the presence of unique OMOP CDM concepts recorded before a patient's index date.
These concepts correspond to coded diagnoses, medication orders, medical device usage, encounter types, lab orders and normal/abnormal result flags, note types, and other data elements extracted from the ``condition\_occurrence'', ``procedure\_occurrence'', ``drug\_exposure'', ``device\_exposure'', ``measurement'', ``note'', and ``observation'' tables in the OMOP CDM.
The extraction procedure for these data elements is repeated separately in three time intervals corresponding to 29 to 1 days prior to the index date, 365 days to 30 days prior to the index, and any time prior to the index date.
Time-agnostic demographic features corresponding to the OMOP CDM concepts for race, ethnicity, and sex are included, as well as a variable indicating the age of the patient at the index date, discretized into five year intervals.
The final feature set is the result of the concatenation of the features derived from each of the described procedures.

For the cohorts derived from MIMIC-III and eICU, we apply the feature extraction code accompanying \citet{harutyunyan2019multitask} and \citet{sheikhalishahi2020benchmarking} to extract demographics and a time-series representation of labs results and vital signs binned into one hour intervals.
Categorical features are one-hot-encoded and numeric features are normalized to zero mean and unit variance.
To the features extracted from MIMIC-III, we include sex as an additional categorical feature and age as an additional numeric feature.
For these cohorts, we evaluate a GRU that operates over a temporal representation, as well as a flattened representation where temporal  numeric features are averaged in 12-hour intervals as inputs to feedforward-neural networks and logistic regression models.

\subsection{Experiments}
\label{sec:experiments}

\subsubsection{Data partitioning}
We partition each cohort such that 62.5\% is used as a training set, 12.5\% is used as a validation set, and 25\% of the data is used as a test set.
Subsequently, the training data is partitioned into five equally-sized folds to enable a modified cross-validation procedure. 
The procedure is conducted for each task by training five models for each hyperparameter configuration, holding out one of the folds of the training set for use as a development set to assess early stopping criteria, and performing model selection based on algorithm-specific model selection criteria defined over the average performance of the five models on the validation set.

\subsubsection{Training and model selection}
\label{sec:methods_training_model_selection}
We conduct a grid search jointly over model-specific and algorithm-specific hyperparameters. 
For ERM experiments trained on the entire population, we evaluate feedforward neural networks for all prediction tasks and additionally apply GRUs to the tasks derived from the MIMIC-III and eICU databases.
For both feedforward neural networks and GRU models, we evaluate a grid of model-specific hyperparameters that includes learning rates of $1 \times 10^{-4}$ and $1 \times 10^{-5}$, one and three hidden layers of size 128 or 256, and a dropout probability of 0.25 or 0.75.
The training procedure is conducted in a minibatch setting of up to 150 iterations of 100 minibatches of size 512 using the Adam \cite{kingma2014adam} optimizer in the Pytorch framework \cite{pytorch}. 
We use early-stopping rules that return the best-performing model seen thus far during training based on criteria applied to the development set when that criteria has not improved for twenty-five epochs of 100 minibatches.
For each combination of model-specific hyperparameters, we evaluate three early stopping criteria that assess either the population average loss, the worst-case subpopulation loss, or the worst-case subpopulation AUC.
We repeat the procedure with a sampling approach that samples an equal proportion of data from each subpopulation in each minibatch.

We conduct a stratified ERM experiment where each of the model-specific hyperparameter configurations assessed in the pooled experiments are applied separately to the data drawn from each subpopulation. 
In addition to the model classes evaluated in other experiments, we also evaluate logistic regression models implemented as zero-layer neural networks with weight decay regularization \cite{loshchilov2018decoupled}. 
We consider weight decay parameters drawn from a grid of values containing 0, 0.01, and 0.001.
For stratified experiments, we use the loss measured on the subpopulation to assess early stopping criteria.

Following training, we apply each model derived from the training procedure to the validation set and assess performance metrics in the pooled population and in each subpopulation.
To select hyperparameters for pooled ERM, we perform selection based on the population average loss.
To evaluate model selection criteria, we compute the average of each resulting performance metric for the set of five models derived from the cross-validation procedure with matching hyperparameters. 
We apply several model selection criteria that mirror the early stopping criteria.
To perform model selection based on the worst-case subpopulation performance, we first compute the average performance of the model replicates on the validation set, for each performance metric and subpopulation. 
Then, we compute the worst-case of the resulting loss or AUC across subpopulations, and take the best worst-case value over all model-specific and algorithm-specific hyperparameters, including early-stopping criteria and sampling rules.
To evaluate the subpopulation balancing approach in isolation, we select the hyperparameter configuration using an average loss across  subpopulations.
Model selection for the stratified ERM experiments occurs based on the average loss over model replicates on the validation set, separately for each subpopulation.

For DRO experiments, we fix model-specific hyperparameters (learning rate, number of hidden layers, size of hidden layers, and dropout probability) to the ones selected for the pooled ERM training procedure. 
We evaluate the five different configurations of DRO outlined in section \ref{sec:dro_methods}.
This consists of the unadjusted formulation of \citet{Sagawa*2020Distributionally}, an adjustment that scales proportionally to the group size, an adjustment that scales inversely to the group size \cite{Sagawa*2020Distributionally}, an adjustment for the marginal entropy of the outcome (the marginal-baselined loss), and the form of the training objective described in section \ref{sec:alternative_metrics} that uses the AUC to steer the optimization process.
For each configuration, we conduct a grid search over hyperparameters including the exponentiated gradient ascent learning rate $\eta$ in the range 1, 0.1, and 0.01, whether to apply subpopulation balancing, and the form of the early stopping rules (either the weighted population loss, implemented as the value of the training objective in equation ($\ref{eq:DRO_theta}$), or the worst-case loss or AUC over subpopulations).
For size-adjusted training objectives, we tune the size adjustment $C$ in the range of 1, 0.1, 0.01. 
For the training objective that uses the marginal-baselined loss, we use stochastic estimates of the marginal entropy using only data from the current minibatch.
For model selection, we extract the hyperparameters with the best worst-case subpopulation performance (both loss and AUC) across all DRO configurations, and separately for each class of DRO training objective.

\subsubsection{Evaluation}
We assess model performance in the test set in terms of AUC, loss, and the absolute calibration error. 
The absolute calibration error assesses the average absolute value of the difference between the outputs of the model and an estimate of the calibration curve constructed via a logistic regression estimator trained on the test data to predict the outcome using the log-transformed outputs of the model as inputs \cite{Pfohl_2021,Austin2019,Yadlowsky2019}.
This formulation is identical to the Integrated Calibration Index of \citet{Austin2019} except that it uses a logistic regression estimator rather than local regression.
To compute 95\% confidence intervals for model performance metrics, we draw 1,000 bootstrap samples from the test set, stratified by levels of the outcome and subpopulation attribute relevant to the evaluation, compute the performance metrics for the set of five derived models on each bootstrap sample, and take the 2.5\% and 97.5\% empirical quantiles of the resulting distribution that results from pooling over both the models and bootstrap replicates.
We construct analogous confidence intervals for the difference in the model performance relative to pooled ERM by computing the difference in the performance on the same bootstrap sample and taking the 2.5\% and 97.5\% empirical quantiles of the distribution of the differences. 
To construct confidence intervals for the worst-case performance over subpopulations, we extract the worst-case performance for each bootstrap sample.

\subsection{Distributionally robust optimization for supervised learning under subpopulation shift} \label{sec:dro_methods}
We consider a supervised learning setting where a dataset $\mathcal{D} = \{(x_i, y_i, a_i)\}_{i = 1}^N \sim P(X, Y, A)$ is used to learn a predictive model $f_{\theta}(x) : \mathbb{R}^{m} \rightarrow [0,1]$ to estimate $\E[Y \mid X=x] = P(Y = 1 \mid X=x)$, where $X \in \mathcal{X} = \mathbb{R}^m$ designates patient-level features, $Y \in \mathcal{Y} = \{0, 1\}$ is a binary indicator for the occurrence of an outcome, and $A \in \mathcal{A}$ is a discrete attribute that stratifies the population into $K$ disjoint subpopulations, where $\mathcal{D}_{A_k} \sim P(X, Y \mid A=A_k)$ corresponds to the subset of $\mathcal{D}$ corresponding to subpopulation $A_k$. 
The standard learning paradigm of ERM seeks a model $f_{\theta}$ that estimates $\E[Y\mid X=x]$ by minimizing the average cross-entropy loss (the empirical risk) $\mathcal{\ell}$ over the dataset:
\begin{equation}
    \min_{\theta \in \Theta} \sum_{i=1}^N \ell(y_i, f_{\theta}(x_i)).
\end{equation}

The framework of DRO \cite{ben2013robust,duchi2018learning,duchi2020distributionally,Hu2018,Sagawa*2020Distributionally} provides the means to formalize the objective of optimizing for the worst-case performance over a set of pre-defined subpopulations.
The general form of the DRO training objective seeks to minimize the expected loss from a worst-case distribution drawn from an uncertainty set of distributions $\mathcal{Q}$:
\begin{equation} \label{eq:dro_general}
    \min_{\theta \in \Theta} \sup_{Q \in \mathcal{Q}} \E_{(x, y) \sim Q} \ell(y, f_{\theta}(x)).
\end{equation}
In the setting of \textit{subpopulation shift}, when $\mathcal{Q}$ is chosen as the set of distributions that result from a change in the subpopulation composition of the population, \textit{i.e.} a change in the marginal distribution $P(A)$, the inner supremum corresponds to a maximization over a weighted combination of the expected losses over each subpopulation \cite{Sagawa*2020Distributionally,Hu2018} that attains its optimum when all of the weight is placed on the subpopulation with the highest loss.
In this case, the definition of the uncertainty set $\mathcal{Q}$ is given by a mixture over the distributions of the data drawn from each group, $\mathcal{Q} := \big\{\sum_{k=1}^K \lambda_k P(X, Y \mid A=A_k)\big \}$,
where $\lambda_k$ is the $k$-th element of a vector of non-negative weights $\lambda \in \Lambda := \{\sum_{k=1}^K \lambda_k = 1 ; \lambda_k \geq 0\}$ that sum to one.
If we let $\ell_k$ be an estimate of $\E_{P(X, Y \mid A=A_k)} \ell(y, f_{\theta}(x))$ computed on a minibatch of data sampled from $\mathcal{D}_{A_k}$, the associated optimization problem can be rewritten as $\min_{\theta \in \Theta} \max_{\lambda \in \Lambda} \sum_{k=1}^{K} \lambda_k \ell_k$.

\citet{Sagawa*2020Distributionally} proposed a stochastic online algorithm for this setting, called GroupDRO (hereafter referred to as DRO).
This algorithm can be described as alternating between exponentiated gradient ascent on the weights $\lambda$
\begin{equation} \label{eq:DRO_lambda}
    \lambda_k \leftarrow \lambda_k \exp(\eta \ell_k) / \sum_{k=1}^K \exp(\eta \ell_k),
\end{equation}
where $\eta$ is a positive scalar learning rate, and stochastic gradient descent (SGD) on the model parameters $\theta$:
\begin{equation} \label{eq:DRO_theta}
    \theta \leftarrow \theta - \eta \nabla_{\theta} \sum_{k=1}^{K} \lambda_k \ell_k.
\end{equation}

\subsubsection{DRO with additive adjustments}
\label{sec:additive_adjustment}
In practice, DRO may perform poorly due to differences across groups in the rate of overfitting  \cite{Sagawa*2020Distributionally}, differences in the amount of irreducible uncertainty in the outcome given the features \cite{Oren2019}, and differences in the variance of the outcome \cite{Meinshausen2015}.
A heuristic approach that has been proposed \cite{Sagawa*2020Distributionally} to improve the empirical behavior of DRO is to introduce subpopulation-specific additive adjustments $c_k$ to the update on the weights $\lambda$:
\begin{equation} \label{eq:DRO_additive}
    \lambda_k \leftarrow \lambda_k \exp(\eta (\ell_k + c_k)) / \sum_{k=1}^K \exp(\eta (\ell_k + c_k)).
\end{equation}
In our experiments, we evaluate two \textit{size-adjusted} updates that scale with the size of group: one where $c_k=\frac{C}{p_k}$ scales with the reciprocal of the relative size of each group $p_k = \frac{n_k}{N}$, where $n_k$ is the number of samples in group $k$, similar to \citet{Sagawa*2020Distributionally}, and one where $c_k= C \sqrt{n_k / N}$ scales proportionally to the group size, where $C$ is a positive scalar hyperparameter.
In addition, we evaluate an approach where $c_k = \E_{P(Y \mid A=A_k)} \log P(Y \mid A=A_k)$ is chosen to be an estimate of the marginal entropy of the outcome in each subpopulation and can either be estimated as a pre-processing step or in a minibatch setting.
We call this the \textit{marginal-baselined loss}, as it is related to the \textit{baselined loss} approach of \citet{Oren2019} that adjusts based on an estimate of conditional entropy.

\subsubsection{Flexible DRO objectives}
\label{sec:alternative_metrics}
We introduce an approach that can incorporate a notion of model performance other than the average loss to assess relative performance of the model across subpopulations, which may be useful for scenarios in which comparisons of the alternative metric across groups are more contextually meaningful than the comparisons of the average loss or its adjusted variants.
We implement this approach as a modified update to $\lambda$ that leaves the form of the update on $\theta$ unchanged.
For a performance metric $g(\mathcal{D}_{A_k}, f_{\theta})$, the form of the associated update on $\lambda$ is
\begin{equation} \label{eq:DRO_lambda_g}
    \lambda_k \leftarrow \lambda_k \exp(\eta g(\mathcal{D}_{A_k}, f_{\theta})) / \sum_{k=1}^K \exp(\eta g(\mathcal{D}_{A_k}, f_{\theta})),
\end{equation}
and the cross entropy loss is used for the update on $\theta$, following equation ($\ref{eq:DRO_theta}$).

We evaluate an instance of this approach that uses the AUC as an example of such a metric given its frequent use as a measure of the performance of clinical predictive models.
In this context, the objective function can be interpreted as empirical risk minimization from the distribution $Q \in \mathcal{Q}$ with the worst-case subpopulation AUC. 
To plug in the AUC to equation (\ref{eq:DRO_lambda_g}), we define a metric $g_\textrm{AUC} = 1-\textrm{AUC}$ such that the maximal $g$ over subpopulations corresponds to the worst-case AUC over subpopulations:
\begin{equation} \label{eq:g_AUC}
    g_\textrm{AUC}(\mathcal{D}_{A_k}, f_{\theta}) = 1-
    \frac{1}{n_k^{y=1} n_k^{y=0}}
    \sum_{i=1}^{n_k^{y=1}} \sum_{j=1}^{n_k^{y=0}} \mathbbm{1}(f_{\theta}(x_i) > f_{\theta}(x_j)).
\end{equation}

\section{Acknowledgements}
We thank the Stanford Center for Population Health Sciences Data Core, the Stanford School of Medicine Research Office, the Stanford Medicine Research IT team, and the Stanford Research Computing Center for supporting the data and computing infrastructure used in this work.
This work is supported by the National Science Foundation Graduate Research Fellowship Program DGE-1656518, National Heart, Lung, and Blood Institute R01 HL144555, and the Stanford Medicine Program for AI in Healthcare.
Any opinions, findings, and conclusions or recommendations expressed in this material are those of the authors and do not necessarily reflect the views of the funding bodies.
\section{Author contributions}
Design of methodology: SRP, HZ, YX, AF, MG, NHS; Software development: SRP, HZ; Data analysis: SRP; Drafting of initial manuscript: SRP; Revision of manuscript: SRP, HZ, YX, AF, MG, NHS; Project administration: NHS; Funding acquisition: NHS, SRP

\section{Data availability}
The availability of the data used in this work is restricted and subject to data use agreements with the respective data owners.
The Stanford Medicine Research Data Repository is not made publicly available.
MIMIC-III and eICU Collaborative Research Database are publicly available following data use agreements with the respective data owners.

\section{Code availability}
We make all code available at \url{https://github.com/som-shahlab/subpopulation_robustness}.

\section{Competing interests statement}
The authors declare no competing interests.

\bibliographystyle{unsrtnat}
\bibliography{bibliography}

\begin{thebibliography}{82}
\providecommand{\natexlab}[1]{#1}
\providecommand{\url}[1]{\texttt{#1}}
\expandafter\ifx\csname urlstyle\endcsname\relax
  \providecommand{\doi}[1]{doi: #1}\else
  \providecommand{\doi}{doi: \begingroup \urlstyle{rm}\Url}\fi

\bibitem[Rajkomar et~al.(2018)Rajkomar, Hardt, Howell, Corrado, and
  Chin]{rajkomar2018-ji}
Alvin Rajkomar, Michaela Hardt, Michael~D. Howell, Greg Corrado, and
  Marshall~H. Chin.
\newblock {Ensuring Fairness in Machine Learning to Advance Health Equity}.
\newblock \emph{Ann. Intern. Med.}, 169\penalty0 (12):\penalty0 866--872, dec
  2018.
\newblock ISSN 0003-4819.
\newblock \doi{10.7326/M18-1990}.

\bibitem[Chen et~al.(2020)Chen, Pierson, Rose, Joshi, Ferryman, and
  Ghassemi]{chen2020ethical}
Irene~Y Chen, Emma Pierson, Sherri Rose, Shalmali Joshi, Kadija Ferryman, and
  Marzyeh Ghassemi.
\newblock Ethical machine learning in healthcare.
\newblock \emph{Annual Review of Biomedical Data Science}, 4, 2020.

\bibitem[Chen et~al.(2019)Chen, Szolovits, and Ghassemi]{chen2019can}
Irene~Y Chen, Peter Szolovits, and Marzyeh Ghassemi.
\newblock Can ai help reduce disparities in general medical and mental health
  care?
\newblock \emph{AMA journal of ethics}, 21\penalty0 (2):\penalty0 167--179,
  2019.

\bibitem[Coley et~al.(2021)Coley, Johnson, Simon, Cruz, and
  Shortreed]{coley2021racial}
R~Yates Coley, Eric Johnson, Gregory~E Simon, Maricela Cruz, and Susan~M
  Shortreed.
\newblock Racial/ethnic disparities in the performance of prediction models for
  death by suicide after mental health visits.
\newblock \emph{JAMA psychiatry}, 2021.

\bibitem[Seyyed-Kalantari et~al.(2020)Seyyed-Kalantari, Liu, McDermott, Chen,
  and Ghassemi]{Seyyed-Kalantari2020}
Laleh Seyyed-Kalantari, Guanxiong Liu, Matthew McDermott, Irene~Y Chen, and
  Marzyeh Ghassemi.
\newblock Chexclusion: Fairness gaps in deep chest x-ray classifiers.
\newblock In \emph{BIOCOMPUTING 2021: Proceedings of the Pacific Symposium},
  pages 232--243. World Scientific, 2020.

\bibitem[Park et~al.(2021)Park, Hu, Singh, Sylla, Dankwa-Mullan, Koski, and
  Das]{park2021comparison}
Yoonyoung Park, Jianying Hu, Moninder Singh, Issa Sylla, Irene Dankwa-Mullan,
  Eileen Koski, and Amar~K Das.
\newblock Comparison of methods to reduce bias from clinical prediction models
  of postpartum depression.
\newblock \emph{JAMA network open}, 4\penalty0 (4):\penalty0 e213909--e213909,
  2021.

\bibitem[Barda et~al.(2021)Barda, Yona, Rothblum, Greenland, Leibowitz,
  Balicer, Bachmat, and Dagan]{barda2021addressing}
Noam Barda, Gal Yona, Guy~N Rothblum, Philip Greenland, Morton Leibowitz, Ran
  Balicer, Eitan Bachmat, and Noa Dagan.
\newblock Addressing bias in prediction models by improving subpopulation
  calibration.
\newblock \emph{Journal of the American Medical Informatics Association},
  28\penalty0 (3):\penalty0 549--558, 2021.

\bibitem[Pfohl et~al.(2019)Pfohl, Marafino, Coulet, Rodriguez, Palaniappan, and
  Shah]{pfohl2019creating}
Stephen Pfohl, Ben Marafino, Adrien Coulet, Fatima Rodriguez, Latha
  Palaniappan, and Nigam~H Shah.
\newblock Creating fair models of atherosclerotic cardiovascular disease risk.
\newblock In \emph{Proceedings of the 2019 AAAI/ACM Conference on AI, Ethics,
  and Society}, pages 271--278, 2019.

\bibitem[Zink and Rose(2020)]{zink2020fair}
Anna Zink and Sherri Rose.
\newblock Fair regression for health care spending.
\newblock \emph{Biometrics}, 76\penalty0 (3):\penalty0 973--982, 2020.

\bibitem[Obermeyer et~al.(2019)Obermeyer, Powers, Vogeli, and
  Mullainathan]{obermeyer2019dissecting}
Ziad Obermeyer, Brian Powers, Christine Vogeli, and Sendhil Mullainathan.
\newblock Dissecting racial bias in an algorithm used to manage the health of
  populations.
\newblock \emph{Science}, 366\penalty0 (6464):\penalty0 447--453, 2019.

\bibitem[Benjamin(2019)]{benjamin2019assessing}
Ruha Benjamin.
\newblock Assessing risk, automating racism.
\newblock \emph{Science}, 366\penalty0 (6464):\penalty0 421--422, 2019.

\bibitem[Paulus and Kent(2020)]{paulus2020predictably}
Jessica~K Paulus and David~M Kent.
\newblock Predictably unequal: understanding and addressing concerns that
  algorithmic clinical prediction may increase health disparities.
\newblock \emph{NPJ digital medicine}, 3\penalty0 (1):\penalty0 1--8, 2020.

\bibitem[Vyas et~al.(2020)Vyas, Eisenstein, and Jones]{Vyas2020}
Darshali~A. Vyas, Leo~G. Eisenstein, and David~S. Jones.
\newblock {Hidden in Plain Sight — Reconsidering the Use of Race Correction
  in Clinical Algorithms}.
\newblock \emph{New England Journal of Medicine}, 383\penalty0 (9):\penalty0
  NEJMms2004740, jun 2020.
\newblock ISSN 0028-4793.
\newblock \doi{10.1056/NEJMms2004740}.

\bibitem[Jacobs and Wallach(2021)]{jacobs2021measurement}
Abigail~Z Jacobs and Hanna Wallach.
\newblock Measurement and fairness.
\newblock In \emph{Proceedings of the 2021 ACM Conference on Fairness,
  Accountability, and Transparency}, pages 375--385, 2021.

\bibitem[Passi and Barocas(2019)]{Passi2019}
Samir Passi and Solon Barocas.
\newblock {Problem formulation and fairness}.
\newblock In \emph{FAT* 2019 - Proceedings of the 2019 Conference on Fairness,
  Accountability, and Transparency}, pages 39--48. Association for Computing
  Machinery, Inc, jan 2019.
\newblock ISBN 9781450361255.
\newblock \doi{10.1145/3287560.3287567}.

\bibitem[Sendak et~al.(2020)Sendak, Gao, Brajer, and
  Balu]{sendak2020presenting}
Mark~P Sendak, Michael Gao, Nathan Brajer, and Suresh Balu.
\newblock Presenting machine learning model information to clinical end users
  with model facts labels.
\newblock \emph{NPJ digital medicine}, 3\penalty0 (1):\penalty0 1--4, 2020.

\bibitem[Gebru et~al.(2018)Gebru, Morgenstern, Vecchione, Vaughan, Wallach,
  Daum{\'e}~III, and Crawford]{gebru2018datasheets}
Timnit Gebru, Jamie Morgenstern, Briana Vecchione, Jennifer~Wortman Vaughan,
  Hanna Wallach, Hal Daum{\'e}~III, and Kate Crawford.
\newblock Datasheets for datasets.
\newblock \emph{arXiv preprint arXiv:1803.09010}, 2018.

\bibitem[Mitchell et~al.(2019)Mitchell, Wu, Zaldivar, Barnes, Vasserman,
  Hutchinson, Spitzer, Raji, and Gebru]{Mitchell2019}
Margaret Mitchell, Simone Wu, Andrew Zaldivar, Parker Barnes, Lucy Vasserman,
  Ben Hutchinson, Elena Spitzer, Inioluwa~Deborah Raji, and Timnit Gebru.
\newblock Model cards for model reporting.
\newblock In \emph{Proceedings of the conference on fairness, accountability,
  and transparency}, pages 220--229, 2019.

\bibitem[Friedler et~al.(2021)Friedler, Scheidegger, and
  Venkatasubramanian]{sorelle_2021_impossibility}
Sorelle~A. Friedler, Carlos Scheidegger, and Suresh Venkatasubramanian.
\newblock The (im)possibility of fairness: Different value systems require
  different mechanisms for fair decision making.
\newblock \emph{Commun. ACM}, 64\penalty0 (4):\penalty0 136–143, March 2021.
\newblock ISSN 0001-0782.
\newblock \doi{10.1145/3433949}.

\bibitem[Jung et~al.(2021)Jung, Kashyap, Avati, Harman, Shaw, Li, Smith, Shum,
  Javitz, Vetteth, et~al.]{jung2021framework}
Kenneth Jung, Sehj Kashyap, Anand Avati, Stephanie Harman, Heather Shaw, Ron
  Li, Margaret Smith, Kenny Shum, Jacob Javitz, Yohan Vetteth, et~al.
\newblock A framework for making predictive models useful in practice.
\newblock \emph{Journal of the American Medical Informatics Association},
  28\penalty0 (6):\penalty0 1149--1158, 2021.

\bibitem[Hardt et~al.(2016)Hardt, Price, Srebro, and Others]{Hardt2016}
Moritz Hardt, Eric Price, Nathan Nati~Nathan Srebro, and Others.
\newblock {Equality of Opportunity in Supervised Learning}.
\newblock \emph{Advances in Neural Information Processing Systems}, pages
  3315--3323, 2016.
\newblock ISSN 10495258.
\newblock \doi{10.1109/ICCV.2015.169}.

\bibitem[Agarwal et~al.(2018)Agarwal, Beygelzimer, Dudik, Langford, and
  Wallach]{Agarwal2018}
Alekh Agarwal, Alina Beygelzimer, Miroslav Dudik, John Langford, and Hanna
  Wallach.
\newblock A reductions approach to fair classification.
\newblock In Jennifer Dy and Andreas Krause, editors, \emph{Proceedings of the
  35th International Conference on Machine Learning}, volume~80 of
  \emph{Proceedings of Machine Learning Research}, pages 60--69,
  Stockholmsmässan, Stockholm Sweden, 10--15 Jul 2018. PMLR.

\bibitem[Celis et~al.(2018)Celis, Huang, Keswani, and Vishnoi]{Celis2018}
L.~Elisa Celis, Lingxiao Huang, Vijay Keswani, and Nisheeth~K. Vishnoi.
\newblock {Classification with Fairness Constraints: A Meta-Algorithm with
  Provable Guarantees}.
\newblock \emph{Proceedings of the Conference on Fairness, Accountability, and
  Transparency}, pages 319--328, jun 2018.

\bibitem[Zafar et~al.(2019)Zafar, Valera, Gomez-Rodriguez, and
  Gummadi]{Zafar2019-cx}
Muhammad~Bilal Zafar, Isabel Valera, Manuel Gomez-Rodriguez, and Krishna~P
  Gummadi.
\newblock {Fairness Constraints: A Flexible Approach for Fair Classification}.
\newblock \emph{J. Mach. Learn. Res.}, 20\penalty0 (75):\penalty0 1--42, 2019.

\bibitem[Kleinberg et~al.(2016)Kleinberg, Mullainathan, and
  Raghavan]{Kleinberg2016}
Jon Kleinberg, Sendhil Mullainathan, and Manish Raghavan.
\newblock {Inherent Trade-Offs in the Fair Determination of Risk Scores}.
\newblock \emph{arXiv preprint arXiv:1609.05807}, 67:\penalty0 43:1----43:23,
  sep 2016.
\newblock ISSN 17409713.
\newblock \doi{10.1111/j.1740-9713.2017.01012.x}.

\bibitem[Chouldechova(2017)]{Chouldechova2017}
Alexandra Chouldechova.
\newblock {Fair prediction with disparate impact: A study of bias in recidivism
  prediction instruments}.
\newblock \emph{ArXiv e-prints}, 5\penalty0 (2):\penalty0 153--163, feb 2017.
\newblock ISSN 2167-6461.
\newblock \doi{10.1089/big.2016.0047}.

\bibitem[Barocas et~al.(2019)Barocas, Hardt, and
  Narayanan]{barocas-hardt-narayanan}
Solon Barocas, Moritz Hardt, and Arvind Narayanan.
\newblock \emph{{Fairness and Machine Learning}}.
\newblock 2019.
\newblock URL \url{fairmlbook.org}.

\bibitem[Pfohl et~al.(2021)Pfohl, Foryciarz, and Shah]{Pfohl_2021}
Stephen~R. Pfohl, Agata Foryciarz, and Nigam~H. Shah.
\newblock An empirical characterization of fair machine learning for clinical
  risk prediction.
\newblock \emph{Journal of Biomedical Informatics}, 113:\penalty0 103621, 2021.
\newblock ISSN 1532-0464.
\newblock \doi{https://doi.org/10.1016/j.jbi.2020.103621}.

\bibitem[Martinez et~al.(2020)Martinez, Bertran, and
  Sapiro]{martinez2020minimax}
Natalia Martinez, Martin Bertran, and Guillermo Sapiro.
\newblock Minimax pareto fairness: A multi objective perspective.
\newblock In \emph{International Conference on Machine Learning}, pages
  6755--6764. PMLR, 2020.

\bibitem[Liu et~al.(2019)Liu, Simchowitz, and Hardt]{Liu2019}
Lydia~T. Liu, Max Simchowitz, and Moritz Hardt.
\newblock {The Implicit Fairness Criterion of Unconstrained Learning}.
\newblock In Kamalika Chaudhuri and Ruslan Salakhutdinov, editors,
  \emph{Proceedings of the 36th International Conference on Machine Learning},
  volume~97 of \emph{Proceedings of Machine Learning Research}, pages
  4051--4060, Long Beach, California, USA, aug 2019. PMLR.

\bibitem[Liu et~al.(2018)Liu, Dean, Rolf, Simchowitz, and
  Hardt]{liu2018delayed}
Lydia~T Liu, Sarah Dean, Esther Rolf, Max Simchowitz, and Moritz Hardt.
\newblock Delayed impact of fair machine learning.
\newblock In \emph{International Conference on Machine Learning}, pages
  3150--3158. PMLR, 2018.

\bibitem[Hu and Chen(2020)]{hu2020fair}
Lily Hu and Yiling Chen.
\newblock Fair classification and social welfare.
\newblock In \emph{Proceedings of the 2020 Conference on Fairness,
  Accountability, and Transparency}, pages 535--545, 2020.

\bibitem[Fazelpour and Lipton(2020)]{fazelpour2020algorithmic}
Sina Fazelpour and Zachary~C Lipton.
\newblock Algorithmic fairness from a non-ideal perspective.
\newblock In \emph{Proceedings of the AAAI/ACM Conference on AI, Ethics, and
  Society}, pages 57--63, 2020.

\bibitem[Corbett-Davies and Goel(2018)]{Corbett-Davies2018}
Sam Corbett-Davies and Sharad Goel.
\newblock {The Measure and Mismeasure of Fairness: A Critical Review of Fair
  Machine Learning}.
\newblock \emph{arXiv preprint arXiv:1808.00023}, jul 2018.
\newblock ISSN 00036951.
\newblock \doi{10.1063/1.3627170}.

\bibitem[Diana et~al.(2021)Diana, Gill, Kearns, Kenthapadi, and
  Roth]{diana2021minimax}
Emily Diana, Wesley Gill, Michael Kearns, Krishnaram Kenthapadi, and Aaron
  Roth.
\newblock Minimax group fairness: Algorithms and experiments.
\newblock In \emph{Proceedings of the AAAI/ACM Conference on AI, Ethics, and
  Society}, 2021.

\bibitem[Sagawa et~al.(2020)Sagawa, Koh, Hashimoto, and
  Liang]{Sagawa*2020Distributionally}
Shiori Sagawa, Pang~Wei Koh, Tatsunori~B Hashimoto, and Percy Liang.
\newblock {Distributionally Robust Neural Networks}.
\newblock In \emph{International Conference on Learning Representations}, 2020.

\bibitem[Ben-Tal et~al.(2013)Ben-Tal, Den~Hertog, De~Waegenaere, Melenberg, and
  Rennen]{ben2013robust}
Aharon Ben-Tal, Dick Den~Hertog, Anja De~Waegenaere, Bertrand Melenberg, and
  Gijs Rennen.
\newblock Robust solutions of optimization problems affected by uncertain
  probabilities.
\newblock \emph{Management Science}, 59\penalty0 (2):\penalty0 341--357, 2013.

\bibitem[Hu et~al.(2018)Hu, Niu, Sato, and Sugiyama]{Hu2018}
Weihua Hu, Gang Niu, Issei Sato, and Masashi Sugiyama.
\newblock Does distributionally robust supervised learning give robust
  classifiers?
\newblock In \emph{International Conference on Machine Learning}, pages
  2029--2037. PMLR, 2018.

\bibitem[Harutyunyan et~al.(2019)Harutyunyan, Khachatrian, Kale, Ver~Steeg, and
  Galstyan]{harutyunyan2019multitask}
Hrayr Harutyunyan, Hrant Khachatrian, David~C Kale, Greg Ver~Steeg, and Aram
  Galstyan.
\newblock Multitask learning and benchmarking with clinical time series data.
\newblock \emph{Scientific data}, 6\penalty0 (1):\penalty0 1--18, 2019.

\bibitem[Sheikhalishahi et~al.(2020)Sheikhalishahi, Balaraman, and
  Osmani]{sheikhalishahi2020benchmarking}
Seyedmostafa Sheikhalishahi, Vevake Balaraman, and Venet Osmani.
\newblock Benchmarking machine learning models on multi-centre eicu critical
  care dataset.
\newblock \emph{PloS one}, 15\penalty0 (7):\penalty0 e0235424, 2020.

\bibitem[Datta et~al.(2020)Datta, Posada, Olson, Li, Balraj, Mesterhazy,
  Pallas, Desai, Shah, O'Reilly, Balraj, Mesterhazy, Pallas, Desai, and
  Shah]{Datta2020}
Somalee Datta, Jose Posada, Garrick Olson, Wencheng Li, Deepa Balraj,
  Joseph~Joe Mesterhazy, Joseph~Joe Pallas, Priyamvada Desai, Nigam~H Shah,
  Ciaran O'Reilly, Deepa Balraj, Joseph~Joe Mesterhazy, Joseph~Joe Pallas,
  Priyamvada Desai, and Nigam~H Shah.
\newblock {A new paradigm for accelerating clinical data science at Stanford
  Medicine}.
\newblock \emph{arXiv preprint arXiv:2003.10534}, mar 2020.

\bibitem[Johnson et~al.(2016)Johnson, Pollard, Shen, Li-Wei, Feng, Ghassemi,
  Moody, Szolovits, Celi, and Mark]{johnson2016mimic}
Alistair~EW Johnson, Tom~J Pollard, Lu~Shen, H~Lehman Li-Wei, Mengling Feng,
  Mohammad Ghassemi, Benjamin Moody, Peter Szolovits, Leo~Anthony Celi, and
  Roger~G Mark.
\newblock Mimic-iii, a freely accessible critical care database.
\newblock \emph{Scientific data}, 3\penalty0 (1):\penalty0 1--9, 2016.

\bibitem[Pollard et~al.(2018)Pollard, Johnson, Raffa, Celi, Mark, and
  Badawi]{pollard2018eicu}
Tom~J Pollard, Alistair~EW Johnson, Jesse~D Raffa, Leo~A Celi, Roger~G Mark,
  and Omar Badawi.
\newblock The eicu collaborative research database, a freely available
  multi-center database for critical care research.
\newblock \emph{Scientific data}, 5\penalty0 (1):\penalty0 1--13, 2018.

\bibitem[He and Garcia(2009)]{he2009learning}
Haibo He and Edwardo~A Garcia.
\newblock Learning from imbalanced data.
\newblock \emph{IEEE Transactions on knowledge and data engineering},
  21\penalty0 (9):\penalty0 1263--1284, 2009.

\bibitem[Austin and Steyerberg(2019)]{Austin2019}
Peter~C. Austin and Ewout~W. Steyerberg.
\newblock {The Integrated Calibration Index (ICI) and related metrics for
  quantifying the calibration of logistic regression models}.
\newblock \emph{Statistics in Medicine}, 38\penalty0 (21):\penalty0 4051--4065,
  sep 2019.
\newblock ISSN 10970258.
\newblock \doi{10.1002/sim.8281}.

\bibitem[Yadlowsky et~al.(2019)Yadlowsky, Basu, and Tian]{Yadlowsky2019}
Steve Yadlowsky, Sanjay Basu, and Lu~Tian.
\newblock A calibration metric for risk scores with survival data.
\newblock In Finale Doshi-Velez, Jim Fackler, Ken Jung, David Kale, Rajesh
  Ranganath, Byron Wallace, and Jenna Wiens, editors, \emph{Proceedings of the
  4th Machine Learning for Healthcare Conference}, volume 106 of
  \emph{Proceedings of Machine Learning Research}, pages 424--450, Ann Arbor,
  Michigan, 09--10 Aug 2019. PMLR.

\bibitem[Cho et~al.(2014)Cho, van Merrienboer, Çaglar Gülçehre, Bahdanau,
  Bougares, Schwenk, and Bengio]{cho2014_gru}
Kyunghyun Cho, Bart van Merrienboer, Çaglar Gülçehre, Dzmitry Bahdanau,
  Fethi Bougares, Holger Schwenk, and Yoshua Bengio.
\newblock Learning phrase representations using rnn encoder-decoder for
  statistical machine translation.
\newblock In \emph{EMNLP}, pages 1724--1734, 2014.

\bibitem[Rosenfeld et~al.(2021{\natexlab{a}})Rosenfeld, Ravikumar, and
  Risteski]{rosenfeld2021the}
Elan Rosenfeld, Pradeep~Kumar Ravikumar, and Andrej Risteski.
\newblock The risks of invariant risk minimization.
\newblock In \emph{International Conference on Learning Representations},
  2021{\natexlab{a}}.

\bibitem[Rosenfeld et~al.(2021{\natexlab{b}})Rosenfeld, Ravikumar, and
  Risteski]{rosenfeld2021online}
Elan Rosenfeld, Pradeep Ravikumar, and Andrej Risteski.
\newblock An online learning approach to interpolation and extrapolation in
  domain generalization.
\newblock \emph{arXiv preprint arXiv:2102.13128}, 2021{\natexlab{b}}.

\bibitem[Koh et~al.(2021)Koh, Sagawa, Marklund, Xie, Zhang, Balsubramani, Hu,
  Yasunaga, Phillips, Gao, Lee, David, Stavness, Guo, Earnshaw, Haque, Beery,
  Leskovec, Kundaje, Pierson, Levine, Finn, and Liang]{Koh2021}
Pang~Wei Koh, Shiori Sagawa, Henrik Marklund, Sang~Michael Xie, Marvin Zhang,
  Akshay Balsubramani, Weihua Hu, Michihiro Yasunaga, Richard~Lanas Phillips,
  Irena Gao, Tony Lee, Etienne David, Ian Stavness, Wei Guo, Berton Earnshaw,
  Imran Haque, Sara~M Beery, Jure Leskovec, Anshul Kundaje, Emma Pierson,
  Sergey Levine, Chelsea Finn, and Percy Liang.
\newblock Wilds: A benchmark of in-the-wild distribution shifts.
\newblock In Marina Meila and Tong Zhang, editors, \emph{Proceedings of the
  38th International Conference on Machine Learning}, volume 139 of
  \emph{Proceedings of Machine Learning Research}, pages 5637--5664. PMLR,
  18--24 Jul 2021.
\newblock URL \url{http://proceedings.mlr.press/v139/koh21a.html}.

\bibitem[Zhang et~al.(2021)Zhang, Dullerud, Seyyed-Kalantari, Morris, Joshi,
  and Ghassemi]{zhang2021empirical}
Haoran Zhang, Natalie Dullerud, Laleh Seyyed-Kalantari, Quaid Morris, Shalmali
  Joshi, and Marzyeh Ghassemi.
\newblock An empirical framework for domain generalization in clinical
  settings.
\newblock In \emph{Proceedings of the Conference on Health, Inference, and
  Learning}, pages 279--290, 2021.

\bibitem[Gulrajani and Lopez-Paz(2020)]{gulrajani2020search}
Ishaan Gulrajani and David Lopez-Paz.
\newblock In search of lost domain generalization.
\newblock \emph{arXiv preprint arXiv:2007.01434}, 2020.

\bibitem[Taori et~al.(2020)Taori, Dave, Shankar, Carlini, Recht, and
  Schmidt]{taori2020measuring}
Rohan Taori, Achal Dave, Vaishaal Shankar, Nicholas Carlini, Benjamin Recht,
  and Ludwig Schmidt.
\newblock Measuring robustness to natural distribution shifts in image
  classification.
\newblock In \emph{Advances in Neural Information Processing Systems
  (NeurIPS)}, 2020.

\bibitem[McDermott et~al.(2021)McDermott, Nestor, Kim, Zhang, Goldenberg,
  Szolovits, and Ghassemi]{McDermott2021}
Matthew McDermott, Bret Nestor, Evan Kim, Wancong Zhang, Anna Goldenberg, Peter
  Szolovits, and Marzyeh Ghassemi.
\newblock {A comprehensive EHR timeseries pre-training benchmark}.
\newblock In \emph{Proceedings of the Conference on Health, Inference, and
  Learning}, volume~21, pages 257--278, New York, NY, USA, apr 2021. ACM.
\newblock ISBN 9781450383592.
\newblock \doi{10.1145/3450439.3451877}.

\bibitem[Steinberg et~al.(2021)Steinberg, Jung, Fries, Corbin, Pfohl, and
  Shah]{steinberg2021language}
Ethan Steinberg, Ken Jung, Jason~A Fries, Conor~K Corbin, Stephen~R Pfohl, and
  Nigam~H Shah.
\newblock Language models are an effective representation learning technique
  for electronic health record data.
\newblock \emph{Journal of Biomedical Informatics}, 113:\penalty0 103637, 2021.

\bibitem[Oakden-Rayner et~al.(2020)Oakden-Rayner, Dunnmon, Carneiro, and
  R{\'{e}}]{Oakden-Rayner2020-wc}
Luke Oakden-Rayner, Jared Dunnmon, Gustavo Carneiro, and Christopher R{\'{e}}.
\newblock {Hidden Stratification Causes Clinically Meaningful Failures in
  Machine Learning for Medical Imaging}.
\newblock \emph{Proc ACM Conf Health Inference Learn (2020)}, 2020:\penalty0
  151--159, apr 2020.

\bibitem[Sohoni et~al.(2020)Sohoni, Dunnmon, Angus, Gu, and R\'{e}]{Sohoni2020}
Nimit Sohoni, Jared Dunnmon, Geoffrey Angus, Albert Gu, and Christopher R\'{e}.
\newblock No subclass left behind: Fine-grained robustness in coarse-grained
  classification problems.
\newblock In H.~Larochelle, M.~Ranzato, R.~Hadsell, M.~F. Balcan, and H.~Lin,
  editors, \emph{Advances in Neural Information Processing Systems}, volume~33,
  pages 19339--19352. Curran Associates, Inc., 2020.

\bibitem[Lahoti et~al.(2020)Lahoti, Beutel, Chen, Lee, Prost, Thain, Wang, and
  Chi]{Lahoti2020-py}
Preethi Lahoti, Alex Beutel, Jilin Chen, Kang Lee, Flavien Prost, Nithum Thain,
  Xuezhi Wang, and Ed~Chi.
\newblock Fairness without demographics through adversarially reweighted
  learning.
\newblock In H.~Larochelle, M.~Ranzato, R.~Hadsell, M.~F. Balcan, and H.~Lin,
  editors, \emph{Advances in Neural Information Processing Systems}, volume~33,
  pages 728--740. Curran Associates, Inc., 2020.

\bibitem[H{\'{e}}bert-Johnson et~al.(2017)H{\'{e}}bert-Johnson, Kim, Reingold,
  and Rothblum]{pmlr-v80-hebert-johnson18a}
{\'{U}}rsula H{\'{e}}bert-Johnson, Michael~P. Kim, Omer Reingold, and Guy~N.
  Rothblum.
\newblock {Calibration for the (Computationally-Identifiable) Masses}.
\newblock In Jennifer Dy and Andreas Krause, editors, \emph{Proceedings of the
  35th International Conference on Machine Learning}, volume~80 of
  \emph{Proceedings of Machine Learning Research}, pages 1939--1948,
  Stockholmsm{\"{a}}ssan, Stockholm Sweden, 2017. PMLR.

\bibitem[Kim et~al.(2019)Kim, Ghorbani, and Zou]{Kim2018}
Michael~P. Kim, Amirata Ghorbani, and James Zou.
\newblock Multiaccuracy: Black-box post-processing for fairness in
  classification.
\newblock In \emph{Proceedings of the 2019 AAAI/ACM Conference on AI, Ethics,
  and Society}, AIES '19, page 247–254, New York, NY, USA, 2019. Association
  for Computing Machinery.
\newblock ISBN 9781450363242.
\newblock \doi{10.1145/3306618.3314287}.

\bibitem[Kearns et~al.(2018)Kearns, Neel, Roth, and Wu]{kearns2017}
Michael Kearns, Seth Neel, Aaron Roth, and Zhiwei~Steven Wu.
\newblock {Preventing Fairness Gerrymandering: Auditing and Learning for
  Subgroup Fairness}.
\newblock \emph{International Conference on Machine Learning}, pages
  2564--2572, nov 2018.
\newblock ISSN 1938-7228.

\bibitem[Creager et~al.(2021)Creager, Jacobsen, and Zemel]{creager2021}
Elliot Creager, Joern-Henrik Jacobsen, and Richard Zemel.
\newblock Environment inference for invariant learning.
\newblock In Marina Meila and Tong Zhang, editors, \emph{Proceedings of the
  38th International Conference on Machine Learning}, volume 139 of
  \emph{Proceedings of Machine Learning Research}, pages 2189--2200. PMLR,
  18--24 Jul 2021.

\bibitem[Miller et~al.(2021)Miller, Gatys, Futoma, and Fox]{miller2021model}
Andrew~C Miller, Leon~A Gatys, Joseph Futoma, and Emily~B Fox.
\newblock Model-based metrics: Sample-efficient estimates of predictive model
  subpopulation performance.
\newblock \emph{arXiv preprint arXiv:2104.12231}, 2021.

\bibitem[Simoiu et~al.(2017)Simoiu, Corbett-Davies, and
  Goel]{simoiu2017problem}
Camelia Simoiu, Sam Corbett-Davies, and Sharad Goel.
\newblock The problem of infra-marginality in outcome tests for discrimination.
\newblock \emph{The Annals of Applied Statistics}, 11\penalty0 (3):\penalty0
  1193--1216, 2017.

\bibitem[Corbett-Davies et~al.(2017)Corbett-Davies, Pierson, Feller, Goel, and
  Huq]{corbett2017algorithmic}
Sam Corbett-Davies, Emma Pierson, Avi Feller, Sharad Goel, and Aziz Huq.
\newblock Algorithmic decision making and the cost of fairness.
\newblock In \emph{Proceedings of the 23rd acm sigkdd international conference
  on knowledge discovery and data mining}, pages 797--806, 2017.

\bibitem[Bakalar et~al.(2021)Bakalar, Barreto, Bergman, Bogen, Chern,
  Corbett-Davies, Hall, Kloumann, Lam, Candela, et~al.]{bakalar2021fairness}
Chlo{\'e} Bakalar, Renata Barreto, Stevie Bergman, Miranda Bogen, Bobbie Chern,
  Sam Corbett-Davies, Melissa Hall, Isabel Kloumann, Michelle Lam,
  Joaquin~Qui{\~n}onero Candela, et~al.
\newblock Fairness on the ground: Applying algorithmic fairness approaches to
  production systems.
\newblock \emph{arXiv preprint arXiv:2103.06172}, 2021.

\bibitem[Foryciarz et~al.(2021)Foryciarz, Pfohl, Patel, and
  Shah]{foryciarz2021evaluating}
Agata Foryciarz, Stephen~R Pfohl, Birju Patel, and Nigam~H Shah.
\newblock Evaluating algorithmic fairness in the presence of clinical
  guidelines: the case of atherosclerotic cardiovascular disease risk
  estimation.
\newblock \emph{medRxiv}, 2021.

\bibitem[Kallus and Zhou(2019)]{kallus2019fairness}
Nathan Kallus and Angela Zhou.
\newblock The fairness of risk scores beyond classification: Bipartite ranking
  and the xauc metric.
\newblock \emph{Advances in neural information processing systems},
  32:\penalty0 3438--3448, 2019.

\bibitem[Narasimhan et~al.(2020)Narasimhan, Cotter, Gupta, and
  Wang]{narasimhan2020pairwise}
Harikrishna Narasimhan, Andrew Cotter, Maya Gupta, and Serena Wang.
\newblock Pairwise fairness for ranking and regression.
\newblock \emph{Proceedings of the AAAI Conference on Artificial Intelligence},
  34:\penalty0 5248--5255, 2020.

\bibitem[Cotter et~al.(2019)Cotter, Jiang, Gupta, Wang, Narayan, You,
  Sridharan, Gupta, You, and Sridharan]{cotter2019optimization}
Andrew Cotter, Heinrich Jiang, Maya~R Gupta, Serena Wang, Taman Narayan,
  Seungil You, Karthik Sridharan, Maya~R Gupta, Seungil You, and Karthik
  Sridharan.
\newblock {Optimization with Non-Differentiable Constraints with Applications
  to Fairness, Recall, Churn, and Other Goals.}
\newblock \emph{Journal of Machine Learning Research}, 20\penalty0
  (172):\penalty0 1--59, sep 2019.

\bibitem[Wald et~al.(2021)Wald, Feder, Greenfeld, and
  Shalit]{wald2021calibration}
Yoav Wald, Amir Feder, Daniel Greenfeld, and Uri Shalit.
\newblock On calibration and out-of-domain generalization.
\newblock \emph{arXiv preprint arXiv:2102.10395}, 2021.

\bibitem[Chen et~al.(2018)Chen, Johansson, and
  Sontag]{ChenJohanssonSontag_NIPS18}
Irene Chen, Fredrik~D. Johansson, and David Sontag.
\newblock {Why Is My Classifier Discriminatory?}
\newblock \emph{Proceedings of the 32nd International Conference on Neural
  Information Processing Systems}, 31:\penalty0 3539--3550, may 2018.

\bibitem[Xu et~al.(2021)Xu, Glicksberg, Su, Walker, Bian, and
  Wang]{xu2021federated}
Jie Xu, Benjamin~S Glicksberg, Chang Su, Peter Walker, Jiang Bian, and Fei
  Wang.
\newblock Federated learning for healthcare informatics.
\newblock \emph{Journal of Healthcare Informatics Research}, 5\penalty0
  (1):\penalty0 1--19, 2021.

\bibitem[Goldberger et~al.(2000)Goldberger, Amaral, Glass, Hausdorff, Ivanov,
  Mark, Mietus, Moody, Peng, and Stanley]{goldberger2000physiobank}
Ary~L Goldberger, Luis~AN Amaral, Leon Glass, Jeffrey~M Hausdorff, Plamen~Ch
  Ivanov, Roger~G Mark, Joseph~E Mietus, George~B Moody, Chung-Kang Peng, and
  H~Eugene Stanley.
\newblock Physiobank, physiotoolkit, and physionet: components of a new
  research resource for complex physiologic signals.
\newblock \emph{circulation}, 101\penalty0 (23):\penalty0 e215--e220, 2000.

\bibitem[Ulmer et~al.(2009)Ulmer, McFadden, and Nerenz]{Ulmer2009}
Cheryl Ulmer, Bernadette McFadden, and David~R Nerenz.
\newblock \emph{{Race, Ethnicity, and Language Data: Standardization for Health
  Care Quality Improvement}}.
\newblock 2009.
\newblock ISBN 978-0-309-14012-6.
\newblock \doi{10.17226/12696}.

\bibitem[Kingma and Ba(2014)]{kingma2014adam}
Diederik~P Kingma and Jimmy Ba.
\newblock Adam: A method for stochastic optimization.
\newblock \emph{arXiv preprint arXiv:1412.6980}, 2014.

\bibitem[Paszke et~al.(2019)Paszke, Gross, Massa, Lerer, Bradbury, Chanan,
  Killeen, Lin, Gimelshein, Antiga, Desmaison, Kopf, Yang, DeVito, Raison,
  Tejani, Chilamkurthy, Steiner, Fang, Bai, and Chintala]{pytorch}
Adam Paszke, Sam Gross, Francisco Massa, Adam Lerer, James Bradbury, Gregory
  Chanan, Trevor Killeen, Zeming Lin, Natalia Gimelshein, Luca Antiga, Alban
  Desmaison, Andreas Kopf, Edward Yang, Zachary DeVito, Martin Raison, Alykhan
  Tejani, Sasank Chilamkurthy, Benoit Steiner, Lu~Fang, Junjie Bai, and Soumith
  Chintala.
\newblock {PyTorch: An Imperative Style, High-Performance Deep Learning
  Library}.
\newblock In H~Wallach, H~Larochelle, A~Beygelzimer, F~d\textquotesingle
  Alch{\'{e}}-Buc, E~Fox, and R~Garnett, editors, \emph{Advances in Neural
  Information Processing Systems 32}, pages 8024--8035. Curran Associates,
  Inc., 2019.

\bibitem[Loshchilov and Hutter(2019)]{loshchilov2018decoupled}
Ilya Loshchilov and Frank Hutter.
\newblock Decoupled weight decay regularization.
\newblock In \emph{International Conference on Learning Representations}, 2019.

\bibitem[Duchi and Namkoong(2018)]{duchi2018learning}
John Duchi and Hongseok Namkoong.
\newblock Learning models with uniform performance via distributionally robust
  optimization.
\newblock \emph{arXiv preprint arXiv:1810.08750}, 2018.

\bibitem[Duchi et~al.(2020)Duchi, Hashimoto, and
  Namkoong]{duchi2020distributionally}
John Duchi, Tatsunori Hashimoto, and Hongseok Namkoong.
\newblock Distributionally robust losses for latent covariate mixtures.
\newblock \emph{arXiv preprint arXiv:2007.13982}, 2020.

\bibitem[Oren et~al.(2019)Oren, Sagawa, Hashimoto, and Liang]{Oren2019}
Yonatan Oren, Shiori Sagawa, Tatsunori~B. Hashimoto, and Percy Liang.
\newblock {Distributionally Robust Language Modeling}.
\newblock \emph{EMNLP-IJCNLP 2019 - 2019 Conference on Empirical Methods in
  Natural Language Processing and 9th International Joint Conference on Natural
  Language Processing, Proceedings of the Conference}, pages 4227--4237, sep
  2019.

\bibitem[Meinshausen et~al.(2015)Meinshausen, B{\"{u}}hlmann, and
  Z{\"{u}}rich]{Meinshausen2015}
Nicolai Meinshausen, Peter B{\"{u}}hlmann, and Eth Z{\"{u}}rich.
\newblock {MAXIMIN EFFECTS IN INHOMOGENEOUS LARGE-SCALE DATA}.
\newblock \emph{The Annals of Statistics}, 43\penalty0 (4):\penalty0
  1801--1830, 2015.
\newblock \doi{10.1214/15-AOS1325}.

\end{thebibliography}

\appendix
\setcounter{page}{1}
\counterwithin{figure}{section}
\counterwithin{table}{section}
\renewcommand*\thetable{\Alph{section}\arabic{table}}
\renewcommand*\thefigure{\Alph{section}\arabic{figure}}
\renewcommand{\figurename}{Supplementary Figure}

\clearpage
\section{Supplementary Tables}

\begin{table}[!th]
\centering
\caption{Characteristics for cohorts drawn from MIMIC-III and the eICU Collaborative Research Database to predict in-hospital mortality 48 hours after ICU admission, following \citet{harutyunyan2019multitask} and \citet{sheikhalishahi2020benchmarking}. 
Data are grouped based on age, sex, and the race and ethnicity category.
Shown, for each group, is the number of patients extracted and the incidence of in-hospital mortality.
}
\label{tab:cohort_mimic_eicu}
\begin{tabular}{lrrrr}
\toprule
{} & \multicolumn{2}{c}{MIMIC-III \cite{harutyunyan2019multitask}} & \multicolumn{2}{c}{eICU \cite{sheikhalishahi2020benchmarking}} \\
\cmidrule{2-5}
Group &   Count & In-hospital mortality &   Count & In-hospital mortality \\
\midrule
\lbrack18-30) &     873 &                 0.056 &   1,301 &                 0.073 \\
\lbrack30-45) &   1,890 &                 0.086 &   2,578 &                 0.074 \\
\lbrack45-55) &   2,916 &                 0.097 &   4,038 &                 0.090 \\
\lbrack55-65) &   4,047 &                 0.109 &   6,458 &                 0.105 \\
\lbrack65-75) &   4,410 &                 0.130 &   7,311 &                 0.116 \\
\lbrack75-90) &   7,003 &                 0.184 &   8,994 &                 0.150 \\
\midrule
Female  &   9,510 &                 0.135 &  13,929 &                 0.116 \\
Male    &  11,629 &                 0.130 &  16,751 &                 0.114 \\
\midrule
Black or African American   &   2,015 &                 0.092 &   3,402 &                 0.096 \\
Other race/ethnicity   &   4,129 &                 0.163 &   3,623 &                 0.114 \\
White   &  14,995 &                 0.129 &  23,655 &                 0.118 \\
\bottomrule
\end{tabular}
\end{table}

\begin{table}[!th]
\centering
\caption{Disaggregated cohort characteristics for patients drawn from STARR included in the ``Other race/ethnicity'' group. Shown, for each group, is the number of patients extracted and the incidence of hospital mortality, prolonged length of stay, and 30-day readmission.}
\label{tab:cohort_starr_admissions_other_race_eth}
\begin{tabular}{lrrrr}
\toprule
{} & {} & \multicolumn{3}{c}{Outcome Incidence} \\
\cmidrule{3-5}
Group &   Count &  Mortality &  Prolonged LOS &  30-day Readmission \\
\midrule
American Indian or Alaska Native          &     502 &              0.0259 &                     0.227 &              0.0558 \\
Native Hawaiian and Pacific Islander &   2,407 &              0.0237 &                      0.210 &              0.0656 \\
Other or no matching race/ethnicity  &  14,789 &              0.0226 &                       0.200 &              0.0449 \\
Patient declined or refused to state      &   1,587 &             0.00945 &                     0.121 &              0.0246 \\
Unknown race/ethnicity               &   5,348 &               0.0570 &                      0.240 &              0.0323 \\
\bottomrule
\end{tabular}
\end{table}

\begin{table}[!th]
\centering
\caption{Disaggregated cohort characteristics for MIMIC-III and the eICU Collaborative Research Database to predict in-hospital mortality 48 hours after ICU admission that were included in the ``Other race/ethnicity'' group. Shown, for each group, is the number of patients extracted and the incidence of in-hospital mortality.}
\label{tab:cohort_mimic_eicu_other_race_eth}
\begin{tabular}{lrrrr}
\toprule
{} & \multicolumn{2}{c}{MIMIC-III \cite{harutyunyan2019multitask}} & \multicolumn{2}{c}{eICU \cite{sheikhalishahi2020benchmarking}} \\
\cmidrule{2-5}
Group &   Count & In-hospital mortality &   Count & In-hospital mortality \\
\midrule
Asian                                &       492 &              0.138 &       492 &              0.118 \\
Hispanic or Latino                   &       679 &              0.0810 &      1,111 &              0.113 \\
Other or unknown race/ethnicity      &          &                    &      1,659 &              0.118 \\
Other or no matching race/ethnicity                 &       573 &              0.140 &          &                    \\
Patient refused or declined to state &       199 &              0.126 &          &                   \\
Unknown race/ethnicity               &      2,186 &              0.203 &          &                    \\
\bottomrule
\end{tabular}
\end{table}

\clearpage
\section{Supplementary Figures}
\begin{figure}[!htb]
    \centering
    \includegraphics[width=0.9\linewidth]{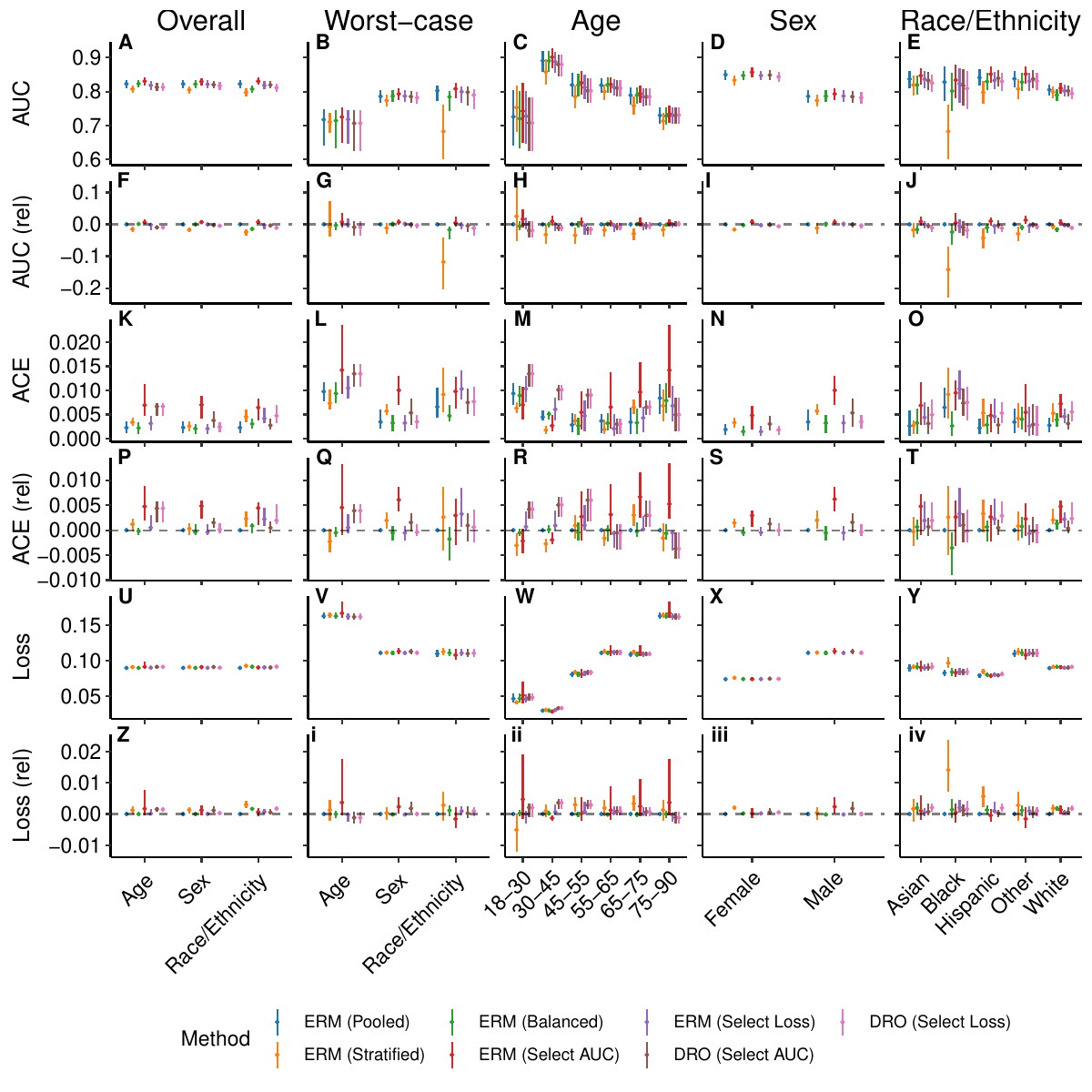}
    \caption{
    The performance of models that to predict in-hospital mortality at admission using data derived from the STARR database. 
    Results shown are the area under the receiver operating characteristic curve (AUC), absolute calibration error (ACE), and the loss assessed in the overall population, on each subpopulation, and in the worst-case over subpopulations for models trained with pooled, stratified, and balanced empirical risk minimization (ERM) and a range of distributionally robust optimization (DRO) training objectives.
    For both pooled ERM and DRO, we show the models selected based on worst-case model selection criteria that perform selection based on the worst-case subpopulation AUC (Select AUC) or loss (Select Loss).
    Model selection occurs over all relevant training objectives, sampling rules, and early-stopping criteria.
    Error bars indicate absolute and relative 95\% confidence intervals derived with the percentile bootstrap with 1,000 iterations.
    Relative performance (suffixed by ``rel'') is assessed with respect to the performance of models derived with ERM applied to the entire training dataset.}
    \label{fig:starr_mortality_performance}
\end{figure}

\begin{figure}[!htb]
    \centering
    \includegraphics[width=0.9\linewidth]{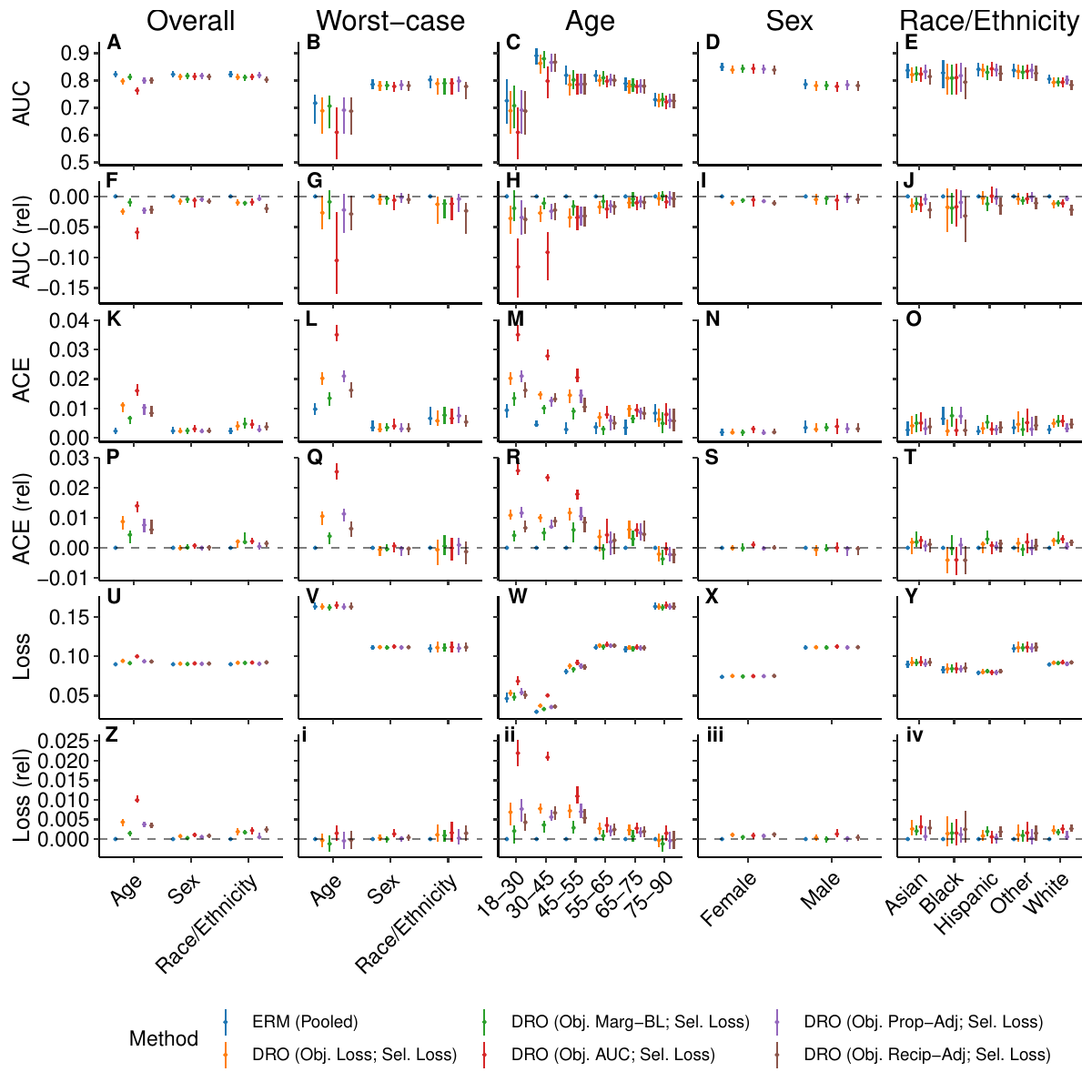}
    \caption{
    The performance of models trained with distributionally robust optimization (DRO) training objectives to predict in-hospital mortality at admission using data derived from the STARR database, following model selection based on worst-case loss over subpopulations. 
    Results shown are the area under the receiver operating characteristic curve (AUC), absolute calibration error (ACE), and the loss assessed in the overall population, on each subpopulation, and in the worst-case over subpopulations for models trained with the unadjusted DRO training objective (Obj. Loss), the adjusted training objective that subtracts the marginal entropy in the outcome (Obj. Marg-BL), the training objective that uses the AUC-based update (Obj. AUC), and training objectives that use adjustments that scale proportionally (Obj. Prop-Adj) and inversely to the size of the group (Obj. Recip-Adj).
    Error bars indicate absolute and relative 95\% confidence intervals derived with the percentile bootstrap with 1,000 iterations.
    Relative performance (suffixed by ``rel'') is assessed with respect to the performance of models derived with ERM applied to the entire training dataset.
    }
    \label{fig:starr_mortality_dro_loss}
\end{figure}

In this section, we provide figures containing the results for each of the prediction tasks evaluated. 
Supplementary Figures \ref{fig:starr_mortality_performance} and \ref{fig:starr_mortality_dro_loss} contain overlapping results with Figures \ref{fig:starr_mortality_performance_relative} and \ref{fig:starr_mortality_dro_loss_relative} presented in main text.
The results for the models learned from the inpatient admission cohort derived from STARR are presented in Section \ref{supp:starr_admissions}.
Section \ref{supp:mimic_eicu_mortality} contains the results for models derived from the MIMIC-III and eICU databases.
\clearpage

\subsection{Results for models that predict in-hospital mortality, prolonged length of stay, and 30-day readmission from STARR}
\label{supp:starr_admissions}

\begin{figure}[!htb]
    \centering
    \includegraphics[width=0.9\linewidth]{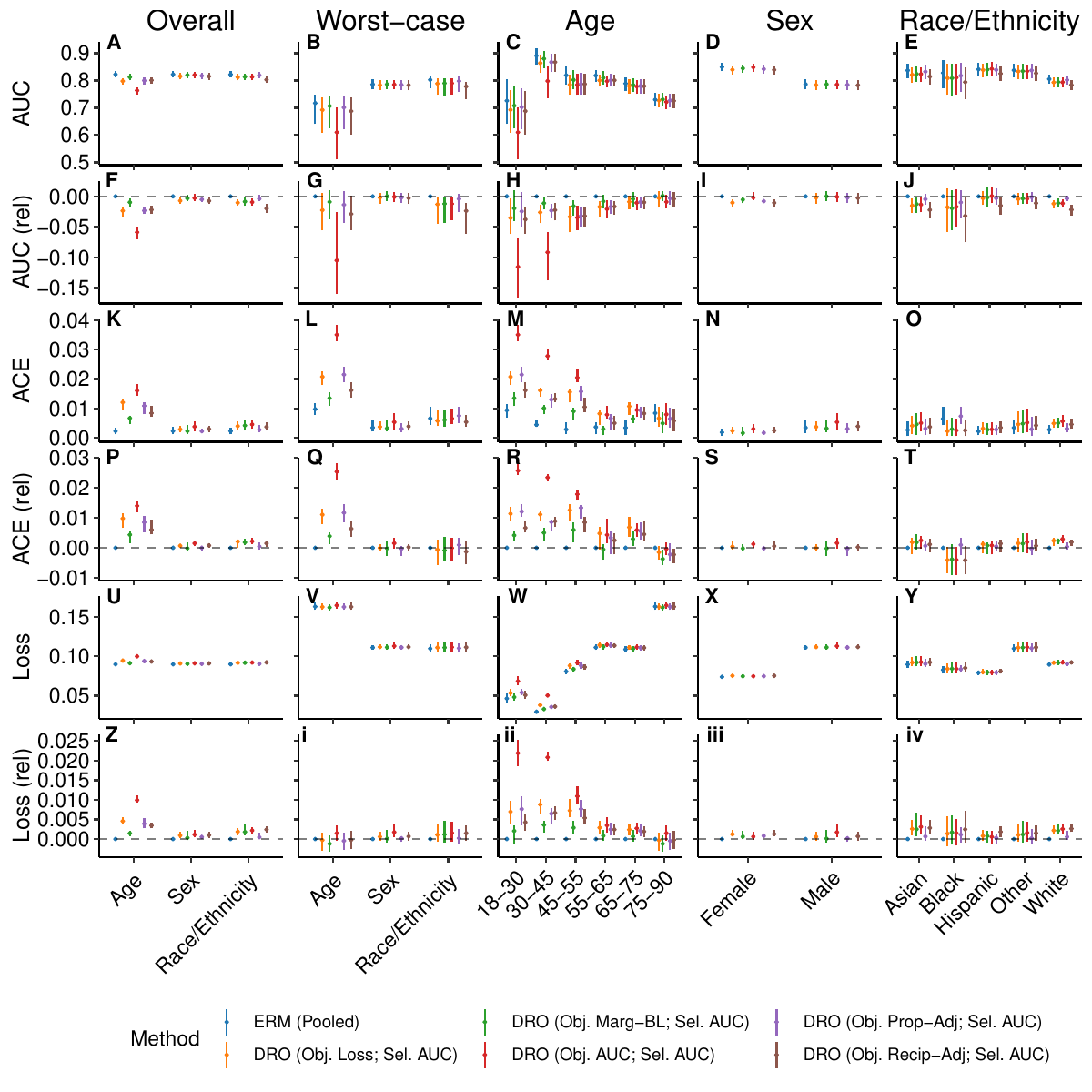}
    \caption{
    The performance of models trained with distributionally robust optimization (DRO) training objectives to predict in-hospital mortality at admission using data derived from the STARR database, following model selection based on worst-case AUC over subpopulations. 
    Results shown are the area under the receiver operating characteristic curve (AUC), absolute calibration error (ACE), and the loss assessed in the overall population, on each subpopulation, and in the worst-case over subpopulations for models trained with the unadjusted DRO training objective (Obj. Loss), the adjusted training objective that subtracts the marginal entropy in the outcome (Obj. Marg-BL), the training objective that uses the AUC-based update (Obj. AUC), and training objectives that use adjustments that scale proportionally (Obj. Prop-Adj) and inversely to the size of the group (Obj. Recip-Adj).
    Error bars indicate absolute and relative 95\% confidence intervals derived with the percentile bootstrap with 1,000 iterations.
    Relative performance (suffixed by ``rel'') is assessed with respect to the performance of models derived with ERM applied to the entire training dataset.
    }
    \label{fig:starr_mortality_dro_auc}
\end{figure}

\begin{figure}[!htb]
    \centering
    \includegraphics[width=0.9\linewidth]{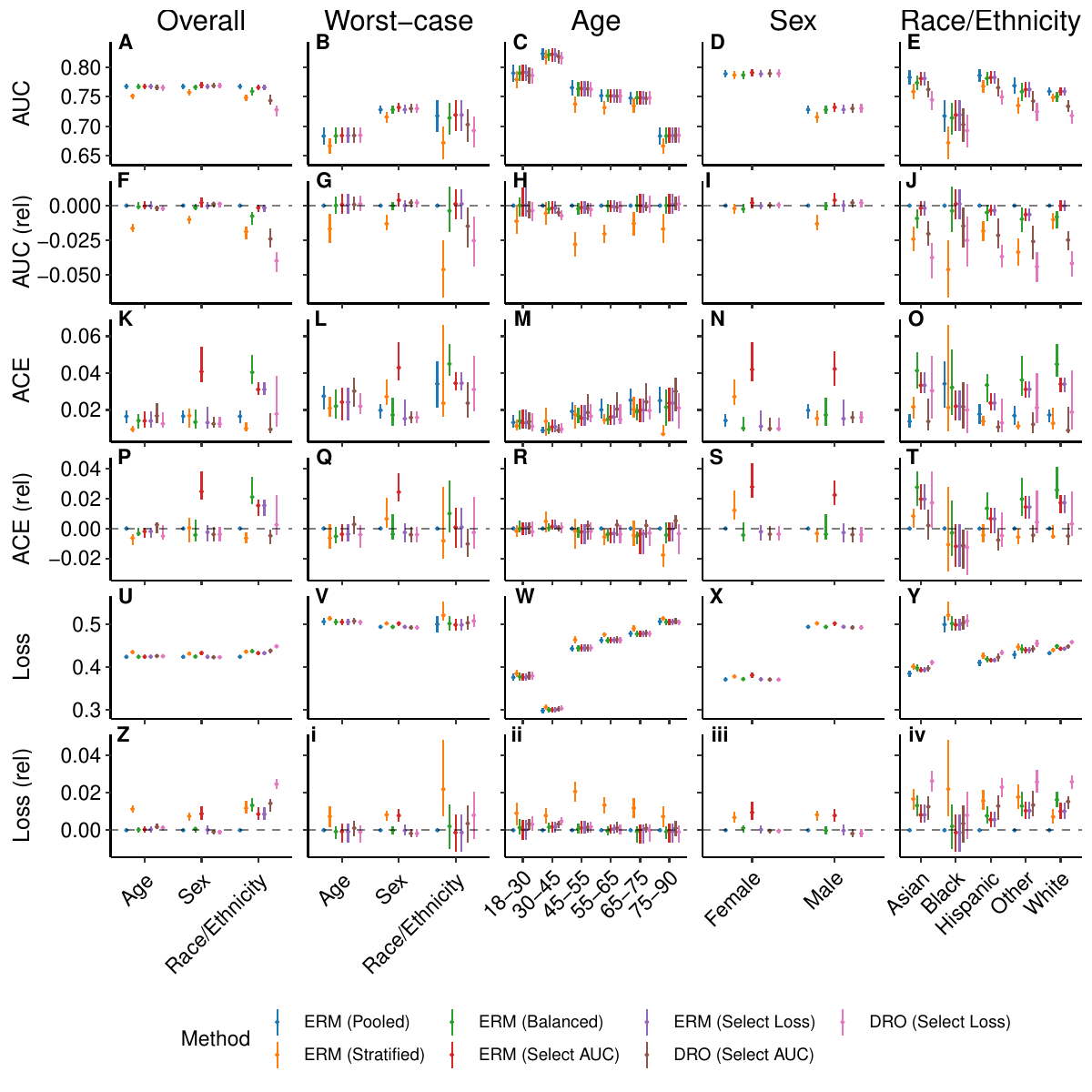}
    \caption{
    The performance of models that predict prolonged length of stay at admission using data derived from the STARR database.
    Results shown are the area under the receiver operating characteristic curve (AUC), absolute calibration error (ACE), and the loss assessed in the overall population, on each subpopulation, and in the worst-case over subpopulations for models trained with pooled, stratified, and balanced empirical risk minimization (ERM) and a range of distributionally robust optimization (DRO) training objectives.
    For both pooled ERM and DRO, we show the models selected based on worst-case model selection criteria that perform selection based on the worst-case subpopulation AUC (Select AUC) or loss (Select Loss).
    Model selection occurs over all relevant training objectives, sampling rules, and early-stopping criteria.
    Error bars indicate absolute and relative 95\% confidence intervals derived with the percentile bootstrap with 1,000 iterations.
    Relative performance (suffixed by ``rel'') is assessed with respect to the performance of models derived with ERM applied to the entire training dataset.
    }
    \label{fig:starr_los_performance}
\end{figure}

\begin{figure}[!htb]
    \centering
    \includegraphics[width=0.9\linewidth]{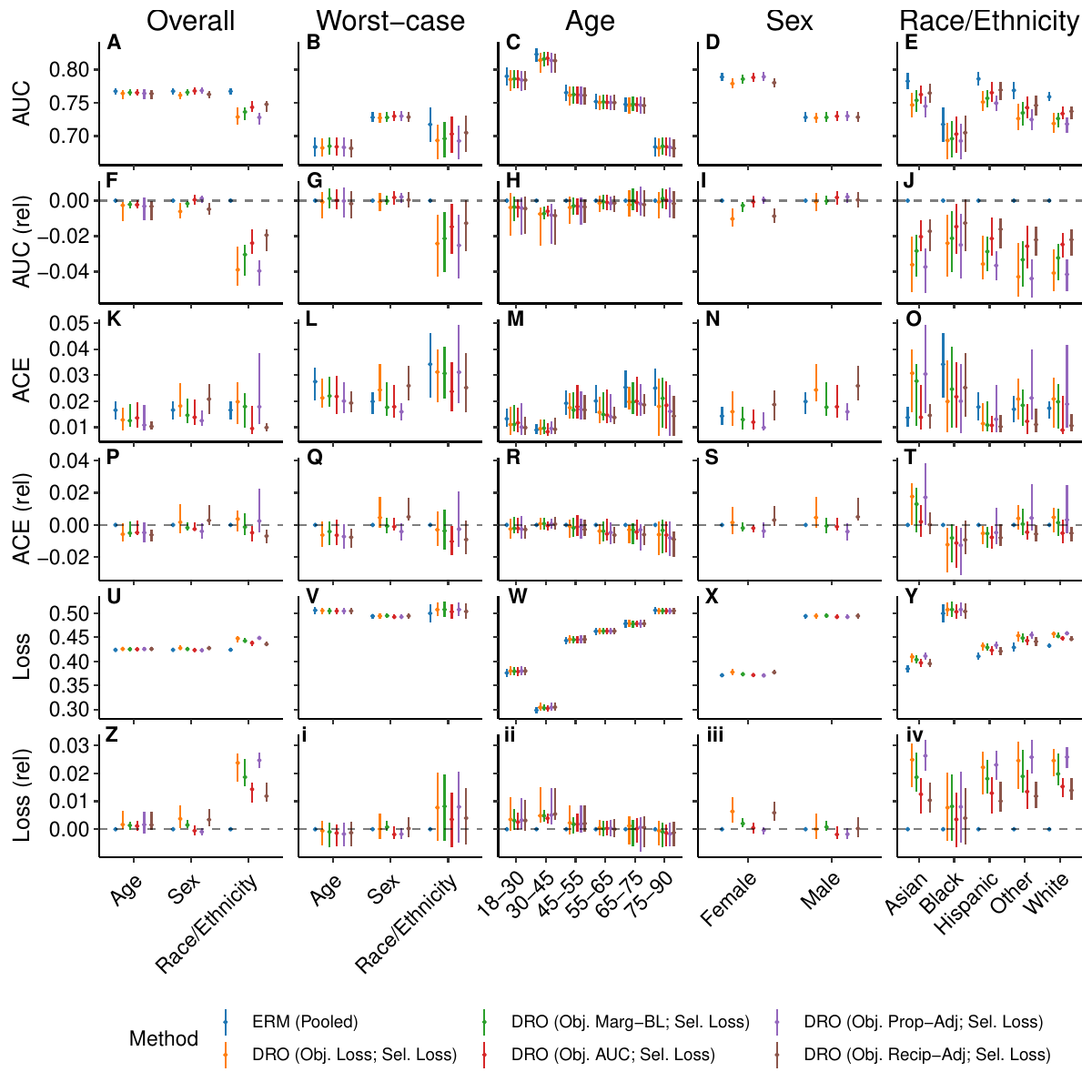}
    \caption{
    The performance of models trained with distributionally robust optimization (DRO) training objectives to predict prolonged length of stay at admission using data derived from the STARR database, following model selection based on worst-case loss over subpopulations. 
    Results shown are the area under the receiver operating characteristic curve (AUC), absolute calibration error (ACE), and the loss assessed in the overall population, on each subpopulation, and in the worst-case over subpopulations for models trained with the unadjusted DRO training objective (Obj. Loss), the adjusted training objective that subtracts the marginal entropy in the outcome (Obj. Marg-BL), the training objective that uses the AUC-based update (Obj. AUC), and training objectives that use adjustments that scale proportionally (Obj. Prop-Adj) and inversely to the size of the group (Obj. Recip-Adj).
    Error bars indicate absolute and relative 95\% confidence intervals derived with the percentile bootstrap with 1,000 iterations.
    Relative performance (suffixed by ``rel'') is assessed with respect to the performance of models derived with ERM applied to the entire training dataset.
    }
    \label{fig:starr_los_dro_loss}
\end{figure}

\begin{figure}[!htb]
    \centering
    \includegraphics[width=0.9\linewidth]{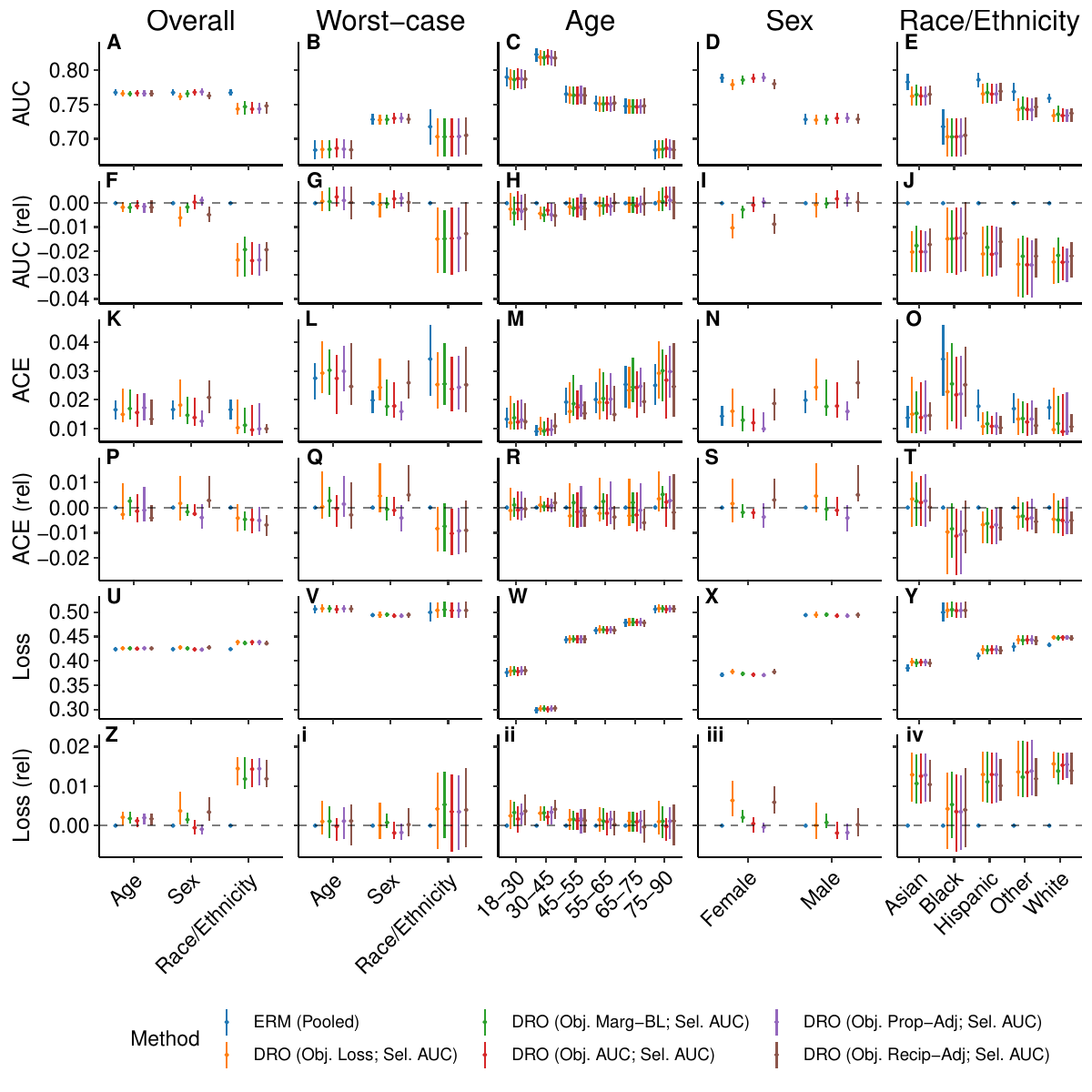}
    \caption{
    The performance of models trained with distributionally robust optimization (DRO) training objectives to predict prolonged length of stay at admission using data derived from the STARR database, following model selection based on worst-case AUC over subpopulations. 
    Results shown are the area under the receiver operating characteristic curve (AUC), absolute calibration error (ACE), and the loss assessed in the overall population, on each subpopulation, and in the worst-case over subpopulations for models trained with the unadjusted DRO training objective (Obj. Loss), the adjusted training objective that subtracts the marginal entropy in the outcome (Obj. Marg-BL), the training objective that uses the AUC-based update (Obj. AUC), and training objectives that use adjustments that scale proportionally (Obj. Prop-Adj) and inversely to the size of the group (Obj. Recip-Adj).
    Error bars indicate absolute and relative 95\% confidence intervals derived with the percentile bootstrap with 1,000 iterations.
    Relative performance (suffixed by ``rel'') is assessed with respect to the performance of models derived with ERM applied to the entire training dataset.
    }
    \label{fig:starr_los_dro_auc}
\end{figure}

\begin{figure}[!htb]
    \centering
    \includegraphics[width=0.9\linewidth]{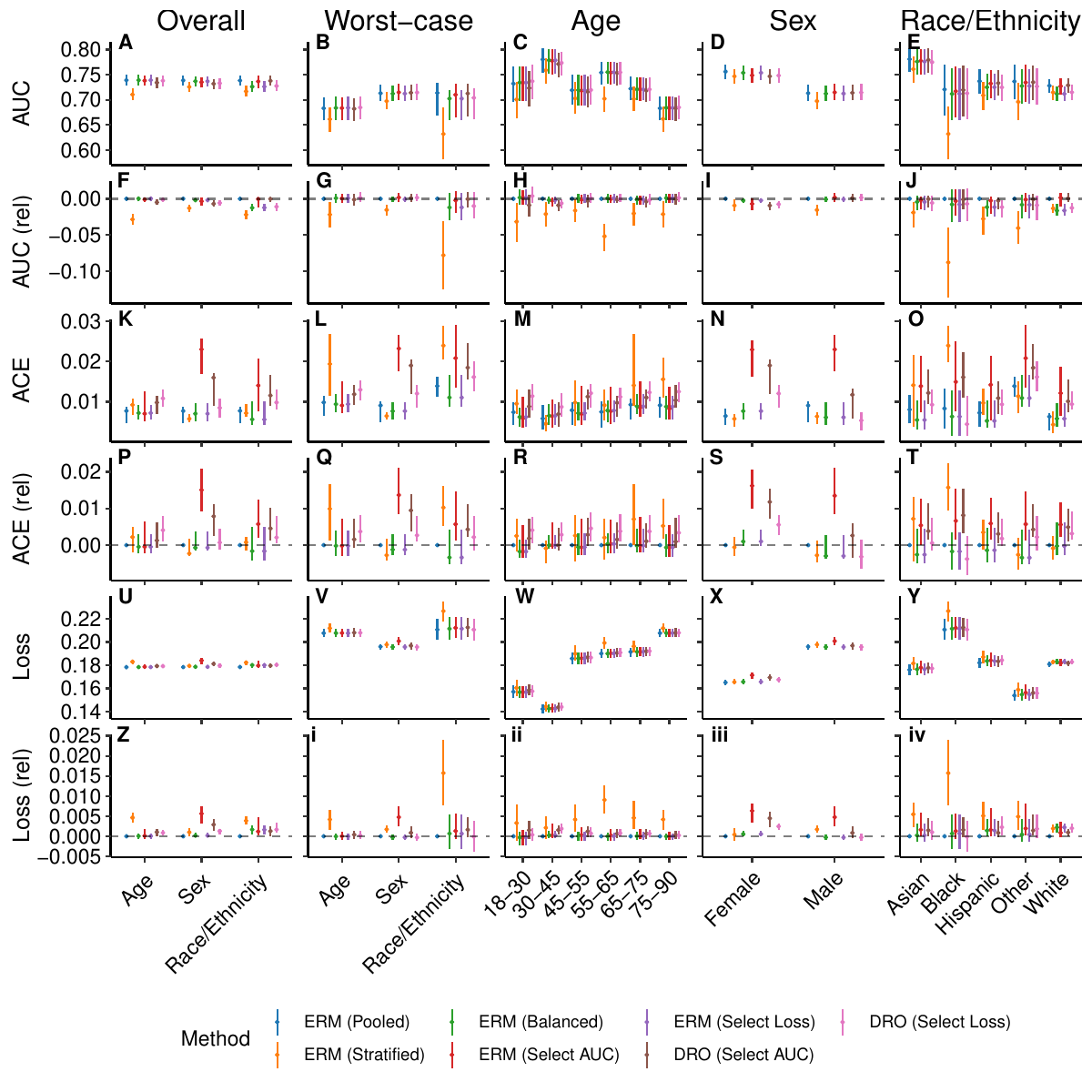}
    \caption{
    The performance of models that predict 30-day readmission at admission using data derived from the STARR database. 
    Results shown are the area under the receiver operating characteristic curve (AUC), absolute calibration error (ACE), and the loss assessed in the overall population, on each subpopulation, and in the worst-case over subpopulations for models trained with pooled, stratified, and balanced empirical risk minimization (ERM) and a range of distributionally robust optimization (DRO) training objectives.
    For both pooled ERM and DRO, we show the models selected based on worst-case model selection criteria that perform selection based on the worst-case subpopulation AUC (Select AUC) or loss (Select Loss).
    Model selection occurs over all relevant training objectives, sampling rules, and early-stopping criteria.
    Error bars indicate absolute and relative 95\% confidence intervals derived with the percentile bootstrap with 1,000 iterations.
    Relative performance (suffixed by ``rel'') is assessed with respect to the performance of models derived with ERM applied to the entire training dataset.
    }
    \label{fig:starr_readmission_performance}
\end{figure}

\begin{figure}[!htb]
    \centering
    \includegraphics[width=0.9\linewidth]{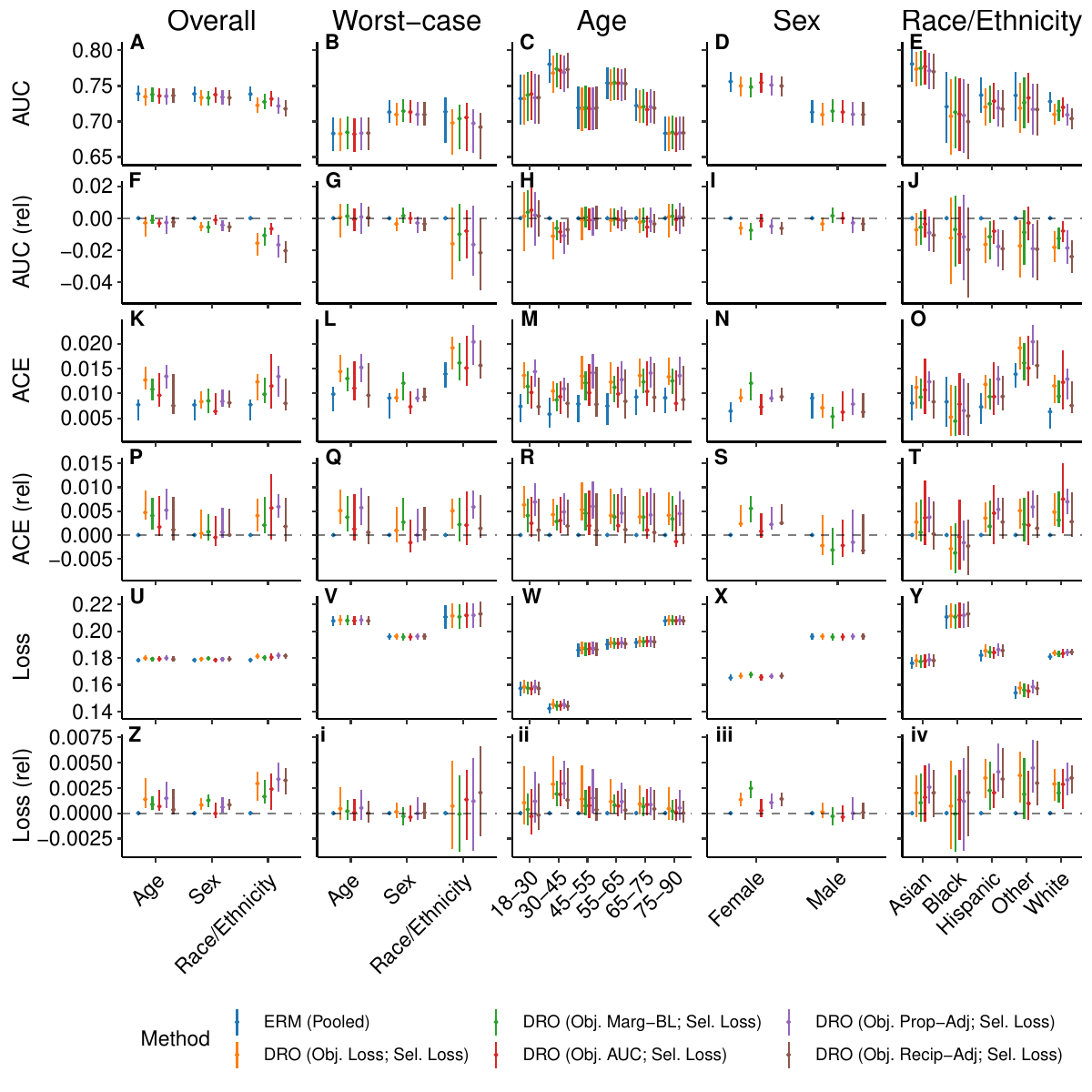}
    \caption{
    The performance of models trained with distributionally robust optimization (DRO) training objectives to predict 30-day readmission at admission using data derived from the STARR database, following model selection based on worst-case loss over subpopulations 
    Results shown are the area under the receiver operating characteristic curve (AUC), absolute calibration error (ACE), and the loss assessed in the overall population, on each subpopulation, and in the worst-case over subpopulations for models trained with the unadjusted DRO training objective (Obj. Loss), the adjusted training objective that subtracts the marginal entropy in the outcome (Obj. Marg-BL), the training objective that uses the AUC-based update (Obj. AUC), and training objectives that use adjustments that scale proportionally (Obj. Prop-Adj) and inversely to the size of the group (Obj. Recip-Adj).
    Error bars indicate absolute and relative 95\% confidence intervals derived with the percentile bootstrap with 1,000 iterations.
    Relative performance (suffixed by ``rel'') is assessed with respect to the performance of models derived with ERM applied to the entire training dataset.
    }
    \label{fig:starr_readmission_dro_loss}
\end{figure}

\begin{figure}[!htb]
    \centering
    \includegraphics[width=0.9\linewidth]{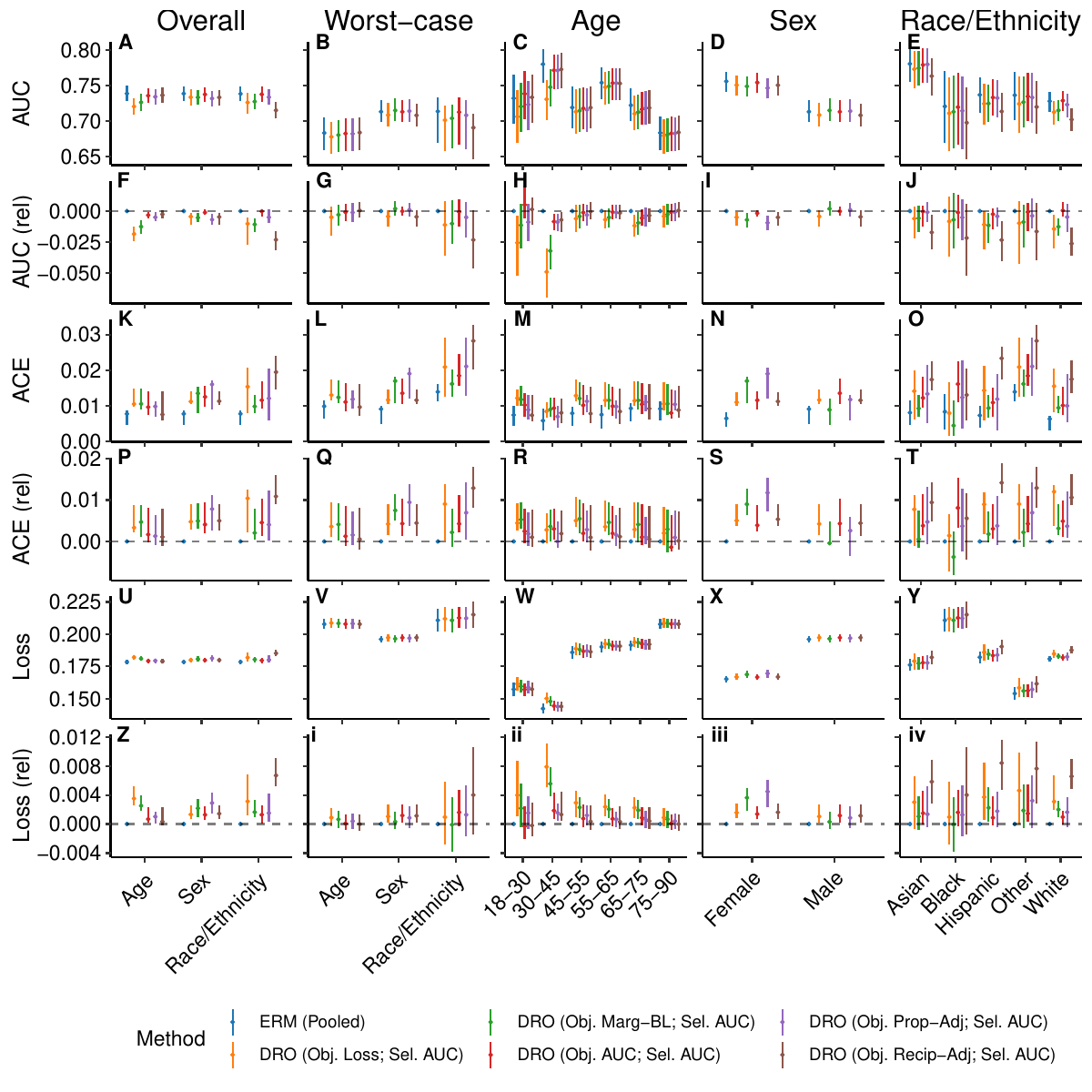}
    \caption{
    The performance of models trained with distributionally robust optimization (DRO) training objectives to predict 30-day readmission at admission using data derived from the STARR database, following model selection based on worst-case AUC over subpopulations. 
    Results shown are the area under the receiver operating characteristic curve (AUC), absolute calibration error (ACE), and the loss assessed in the overall population, on each subpopulation, and in the worst-case over subpopulations for models trained with the unadjusted DRO training objective (Obj. Loss), the adjusted training objective that subtracts the marginal entropy in the outcome (Obj. Marg-BL), the training objective that uses the AUC-based update (Obj. AUC), and training objectives that use adjustments that scale proportionally (Obj. Prop-Adj) and inversely to the size of the group (Obj. Recip-Adj).
    Error bars indicate absolute and relative 95\% confidence intervals derived with the percentile bootstrap with 1,000 iterations.
    Relative performance (suffixed by ``rel'') is assessed with respect to the performance of models derived with ERM applied to the entire training dataset.
    }
    \label{fig:starr_readmission_dro_auc}
\end{figure}

\clearpage
\subsection{Results for models that predict in-hospital mortality from intensive care databases}
\label{supp:mimic_eicu_mortality}
\begin{figure}[!htb]
    \centering
    \includegraphics[width=0.9\linewidth]{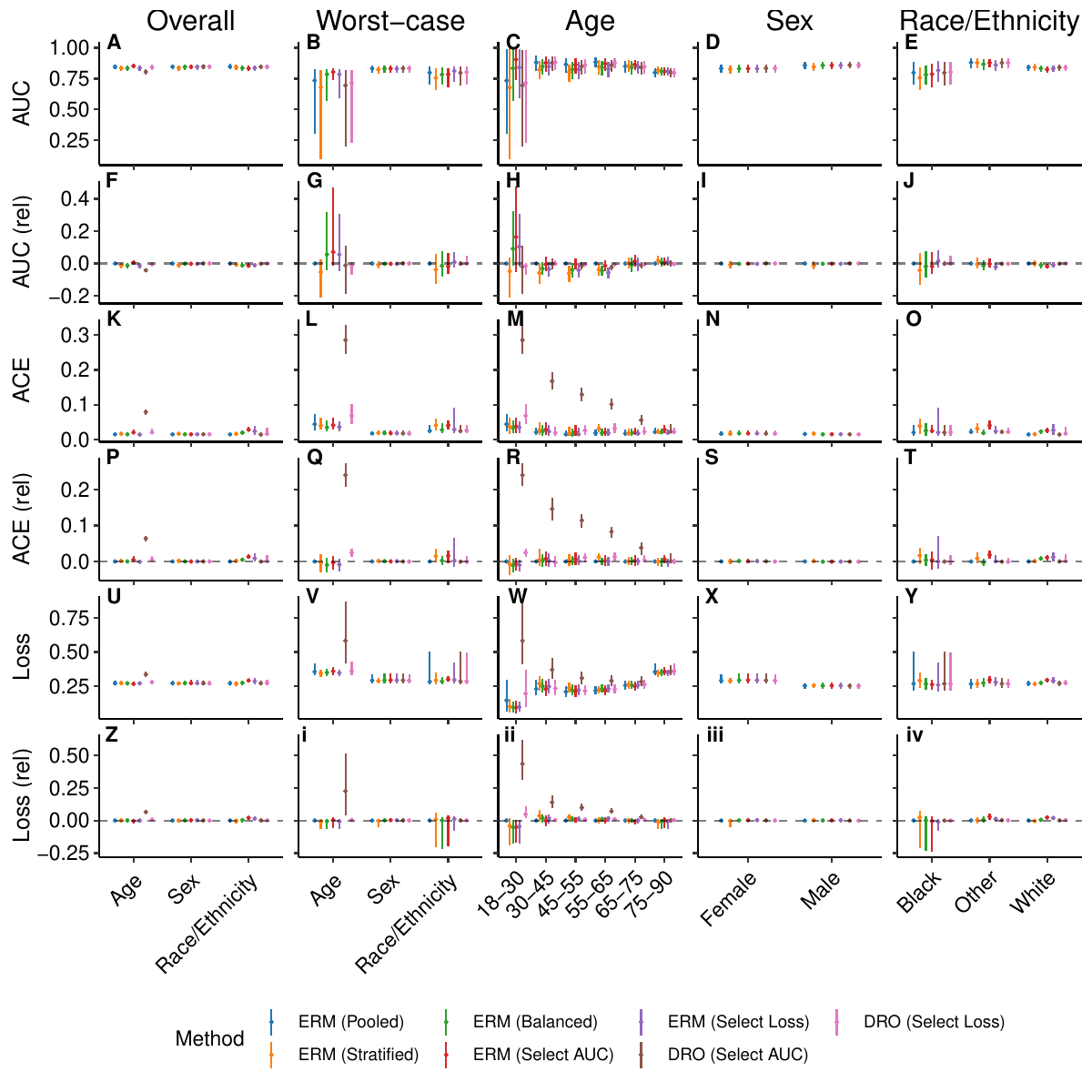}
    \caption{The performance of models that predict in-hospital mortality using features derived from data recorded in the first 48 hours of a patient's ICU stay for data derived from the MIMIC-III database, following \citet{harutyunyan2019multitask}.
    Results shown are the area under the receiver operating characteristic curve (AUC), absolute calibration error (ACE), and the loss assessed in the overall population, on each subpopulation, and in the worst-case over subpopulations for models trained with pooled, stratified, and balanced empirical risk minimization (ERM) and a range of distributionally robust optimization (DRO) training objectives.
    For both pooled ERM and DRO, we show the models selected based on worst-case model selection criteria that perform selection based on the worst-case subpopulation AUC (Select AUC) or loss (Select Loss).
    Model selection occurs over all relevant training objectives, sampling rules, and early-stopping criteria.
    Error bars indicate absolute and relative 95\% confidence intervals derived with the percentile bootstrap with 1,000 iterations.
    Relative performance (suffixed by ``rel'') is assessed with respect to the performance of models derived with ERM applied to the entire training dataset.}
    \label{fig:mimic_mortality_vitals_performance}
\end{figure}

\begin{figure}[!htb]
    \centering
    \includegraphics[width=0.9\linewidth]{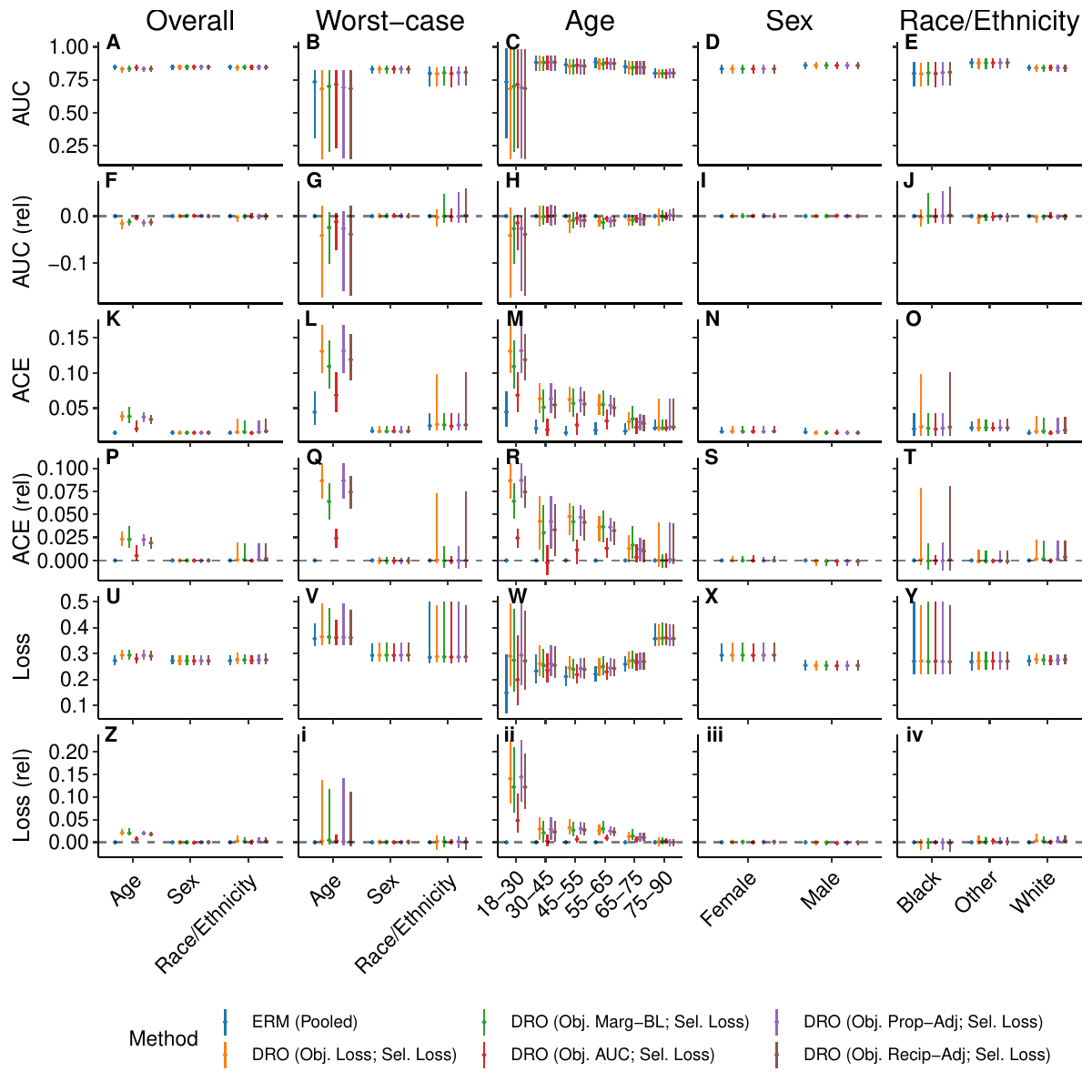}
    \caption{The performance of models trained with distributionally robust optimization (DRO) training objectives to predict in-hospital mortality using features extracted from data derived from the first 48 hours of a patient's ICU stay using data derived from the MIMIC-III database, following \citet{harutyunyan2019multitask}, following model selection based on worst-case loss over subpopulations 
    Results shown are the area under the receiver operating characteristic curve (AUC), absolute calibration error (ACE), and the loss assessed in the overall population, on each subpopulation, and in the worst-case over subpopulations for models trained with the unadjusted DRO training objective (Obj. Loss), the adjusted training objective that subtracts the marginal entropy in the outcome (Obj. Marg-BL), the training objective that uses the AUC-based update (Obj. AUC), and training objectives that use adjustments that scale proportionally (Obj. Prop-Adj) and inversely to the size of the group (Obj. Recip-Adj).
    Error bars indicate absolute and relative 95\% confidence intervals derived with the percentile bootstrap with 1,000 iterations.
    Relative performance (suffixed by ``rel'') is assessed with respect to the performance of models derived with ERM applied to the entire training dataset.}
    \label{fig:mimic_mortality_vitals_dro_loss}
\end{figure}

\begin{figure}[!htb]
    \centering
    \includegraphics[width=0.9\linewidth]{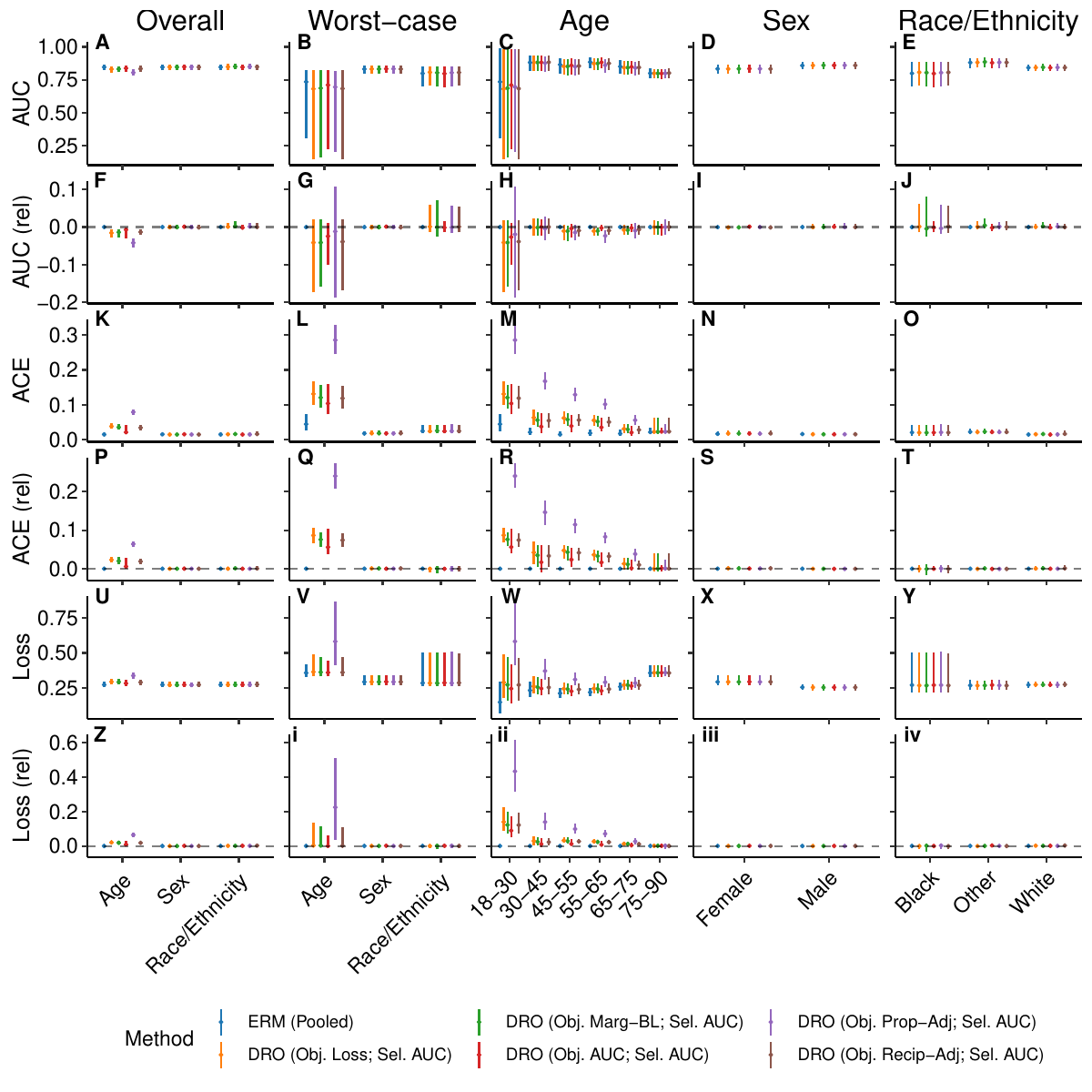}
    \caption{The performance of models trained with distributionally robust optimization (DRO) training objectives to predict in-hospital mortality using features extracted from data derived from the first 48 hours of a patient's ICU stay using data derived from the MIMIC-III database, following \citet{harutyunyan2019multitask}, following model selection based on worst-case AUC over subpopulations. 
    Results shown are the area under the receiver operating characteristic curve (AUC), absolute calibration error (ACE), and the loss assessed in the overall population, on each subpopulation, and in the worst-case over subpopulations for models trained with the unadjusted DRO training objective (Obj. Loss), the adjusted training objective that subtracts the marginal entropy in the outcome (Obj. Marg-BL), the training objective that uses the AUC-based update (Obj. AUC), and training objectives that use adjustments that scale proportionally (Obj. Prop-Adj) and inversely to the size of the group (Obj. Recip-Adj).
    Error bars indicate absolute and relative 95\% confidence intervals derived with the percentile bootstrap with 1,000 iterations.
    Relative performance (suffixed by ``rel'') is assessed with respect to the performance of models derived with ERM applied to the entire training dataset.}
    \label{fig:mimic_mortality_vitals_dro_auc}
\end{figure}

\begin{figure}[!htb]
    \centering
    \includegraphics[width=0.9\linewidth]{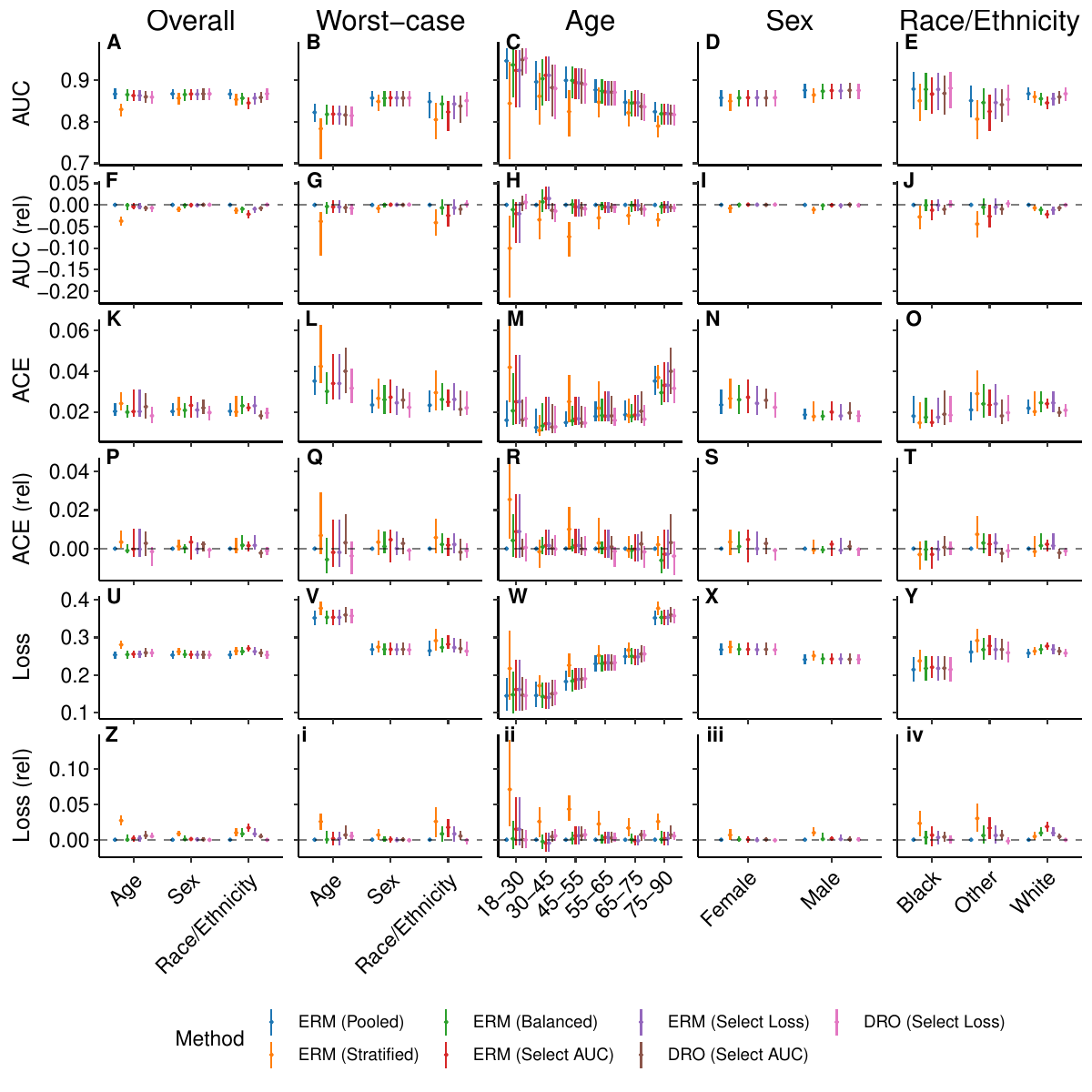}
    \caption{The performance of models that predict in-hospital mortality using features derived from data recorded in the first 48 hours of a patient's ICU stay for data derived from the eICU database, following \citet{sheikhalishahi2020benchmarking}.
    Results shown are the area under the receiver operating characteristic curve (AUC), absolute calibration error (ACE), and the loss assessed in the overall population, on each subpopulation, and in the worst-case over subpopulations for models trained with pooled, stratified, and balanced empirical risk minimization (ERM) and a range of distributionally robust optimization (DRO) training objectives.
    For both pooled ERM and DRO, we show the models selected based on worst-case model selection criteria that perform selection based on the worst-case subpopulation AUC (Select AUC) or loss (Select Loss).
    Model selection occurs over all relevant training objectives, sampling rules, and early-stopping criteria.
    Error bars indicate absolute and relative 95\% confidence intervals derived with the percentile bootstrap with 1,000 iterations.
    Relative performance (suffixed by ``rel'') is assessed with respect to the performance of models derived with ERM applied to the entire training dataset.}
    \label{fig:eicu_mortality_performance}
\end{figure}

\begin{figure}[!htb]
    \centering
    \includegraphics[width=0.9\linewidth]{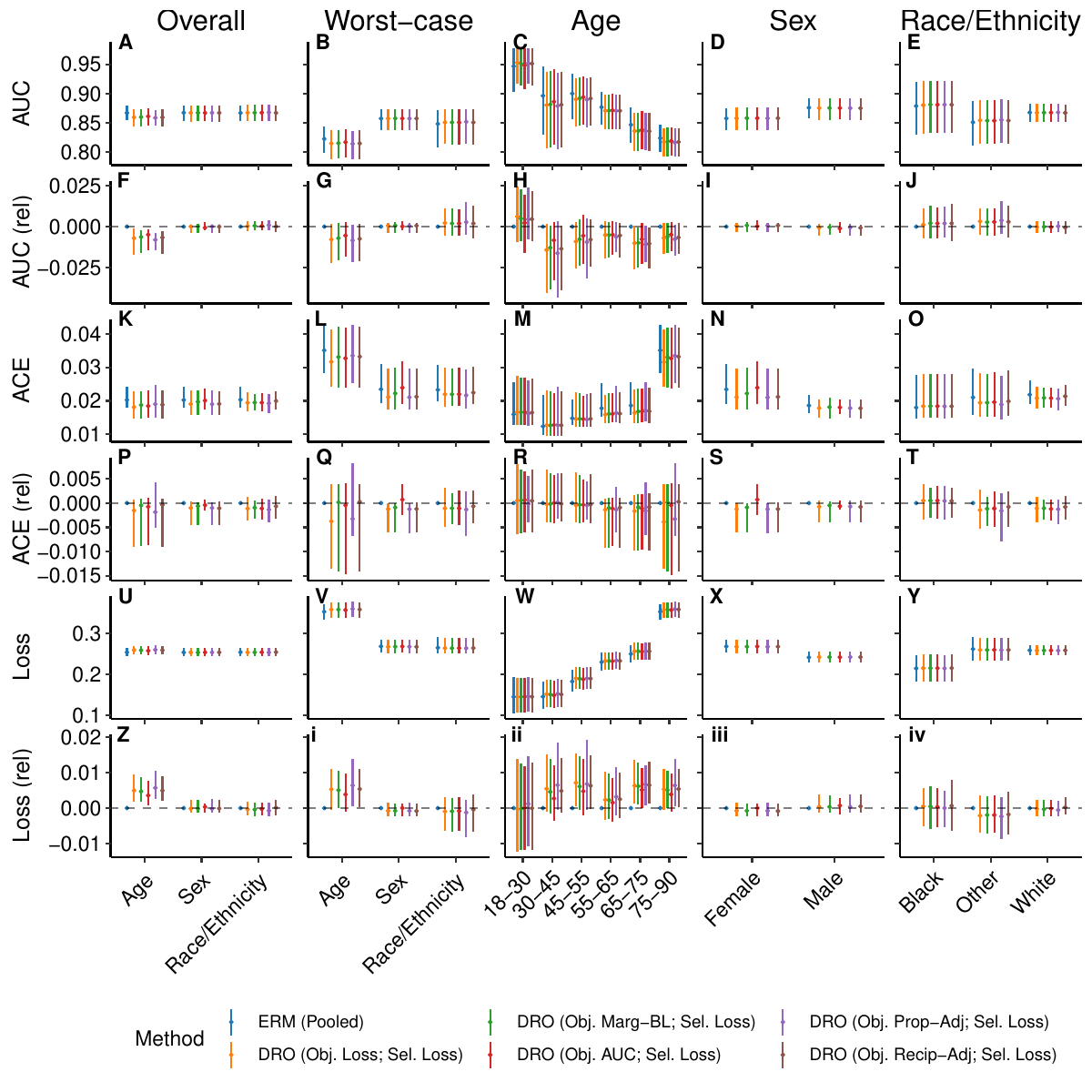}
    \caption{The performance of models trained with distributionally robust optimization (DRO) training objectives to predict in-hospital mortality using features extracted from data derived from the first 48 hours of a patient's ICU stay using data derived from the eICU database, following \citet{sheikhalishahi2020benchmarking}, following model selection based on worst-case loss over subpopulations 
    Results shown are the area under the receiver operating characteristic curve (AUC), absolute calibration error (ACE), and the loss assessed in the overall population, on each subpopulation, and in the worst-case over subpopulations for models trained with the unadjusted DRO training objective (Obj. Loss), the adjusted training objective that subtracts the marginal entropy in the outcome (Obj. Marg-BL), the training objective that uses the AUC-based update (Obj. AUC), and training objectives that use adjustments that scale proportionally (Obj. Prop-Adj) and inversely to the size of the group (Obj. Recip-Adj).
    Error bars indicate absolute and relative 95\% confidence intervals derived with the percentile bootstrap with 1,000 iterations.
    Relative performance (suffixed by ``rel'') is assessed with respect to the performance of models derived with ERM applied to the entire training dataset.}
    \label{fig:eicu_mortality_dro_loss}
\end{figure}

\begin{figure}[!htb]
    \centering
    \includegraphics[width=0.9\linewidth]{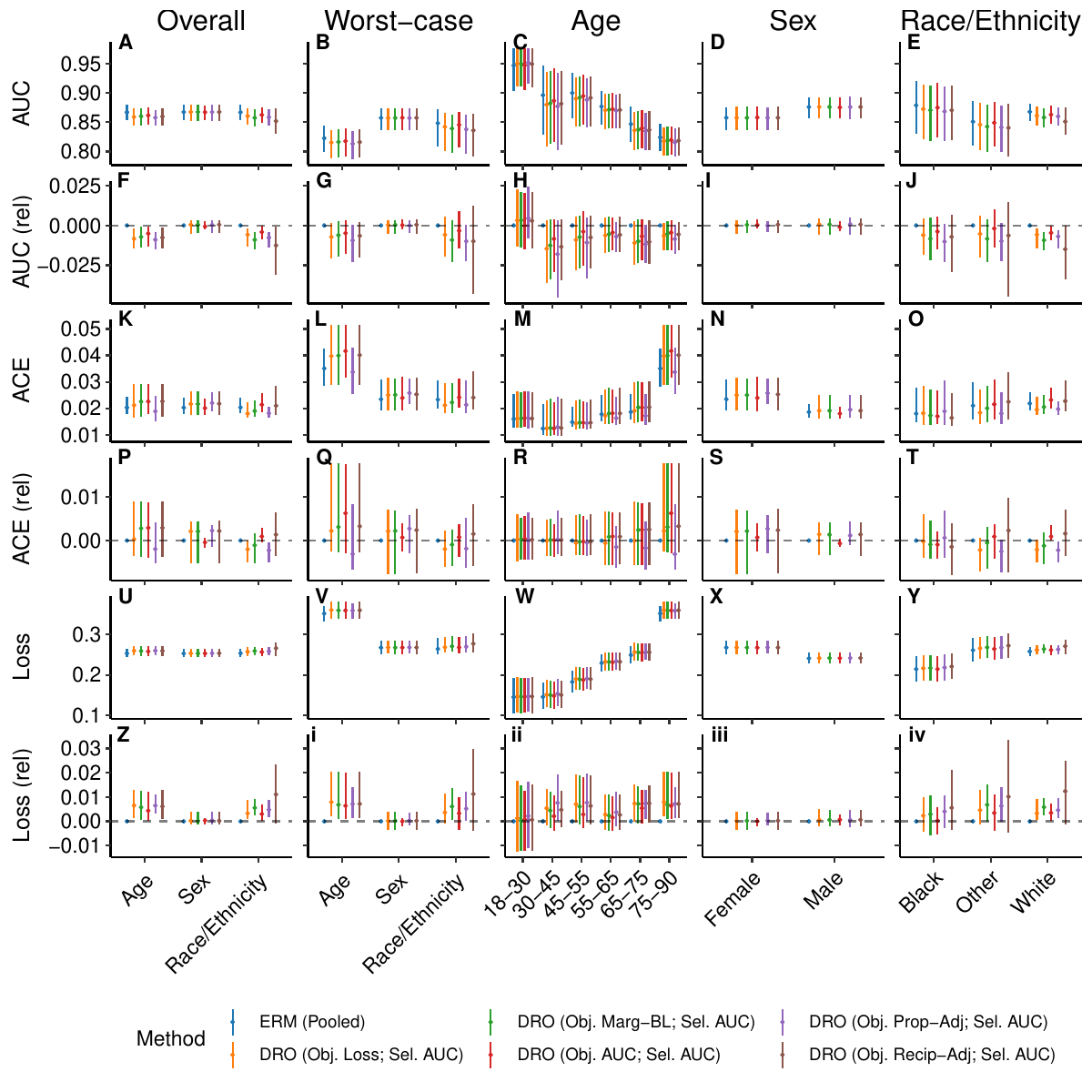}
    \caption{The performance of models trained with distributionally robust optimization (DRO) training objectives to predict in-hospital mortality using features extracted from data derived from the first 48 hours of a patient's ICU stay using data derived from the eICU database, following \citet{sheikhalishahi2020benchmarking}, following model selection based on worst-case AUC over subpopulations. 
    Results shown are the area under the receiver operating characteristic curve (AUC), absolute calibration error (ACE), and the loss assessed in the overall population, on each subpopulation, and in the worst-case over subpopulations for models trained with the unadjusted DRO training objective (Obj. Loss), the adjusted training objective that subtracts the marginal entropy in the outcome (Obj. Marg-BL), the training objective that uses the AUC-based update (Obj. AUC), and training objectives that use adjustments that scale proportionally (Obj. Prop-Adj) and inversely to the size of the group (Obj. Recip-Adj).
    Error bars indicate absolute and relative 95\% confidence intervals derived with the percentile bootstrap with 1,000 iterations.
    Relative performance (suffixed by ``rel'') is assessed with respect to the performance of models derived with ERM applied to the entire training dataset.}
    \label{fig:eicu_mortality_dro_auc}
\end{figure}

\end{document}